\documentclass[acmlarge]{acmart}
\usepackage{amsfonts}
\usepackage{amsmath}
\usepackage[capitalise]{cleveref}
\usepackage{enumitem}
\usepackage{tabu}
\usepackage{soul}
\usepackage{subcaption}
\usepackage{booktabs}
\usepackage{multirow}
\usepackage{wrapfig}

\AtBeginDocument{%
  \providecommand\BibTeX{{%
    \normalfont B\kern-0.5em{\scshape i\kern-0.25em b}\kern-0.8em\TeX}}}

\setcopyright{acmcopyright}
\copyrightyear{2022}
\acmYear{2022}
\acmDOI{XXXXXXX.XXXXXXX}

\acmJournal{POMACS}
\acmVolume{37}
\acmNumber{4}
\acmArticle{111}
\acmMonth{8}

\newcommand{\ie}{\textit{i.e.}}
\newcommand{\eg}{\textit{e.g.}}

\begin{document}

\title{$\textup{D}^{\textup{3}}$T-GAN: Data-Dependent Domain Transfer GANs for Few-shot Image Generation}

\author{Xintian Wu}
\email{hsintien@zju.edu.cn}
\affiliation{
 \institution{Zhejiang University}
	\department{College of Computer Science}
	\city{Hangzhou}
	\state{Zhejiang Province}
	\country{China}
}

\author{Huanyu Wang}
\email{huanyuhello@zju.edu.cn}
\affiliation{
 \institution{Zhejiang University}
	\department{College of Computer Science}
	\city{Hangzhou}
	\state{Zhejiang Province}
	\country{China}
}

\author{Yiming Wu}
\email{ymw@zju.edu.cn}
\affiliation{
 \institution{Zhejiang University}
	\department{College of Computer Science}
	\city{Hangzhou}
	\state{Zhejiang Province}
	\country{China}
}

\author{Xi Li}
\email{xilizju@zju.edu.cn}
\authornote{Corresponding author.}
\affiliation{
 \institution{Zhejiang University}
	\department{College of Computer Science}
	\city{Hangzhou}
	\state{Zhejiang Province}
	\country{China}
}

\renewcommand{\shortauthors}{Wu et al.}

\begin{abstract}
	
	As an important and challenging problem, few-shot image generation aims at generating realistic images through training a GAN model given few samples. A typical solution for few-shot generation is to transfer a well-trained GAN model from a data-rich source domain to the data-deficient target domain. In this paper, we propose a novel self-supervised transfer scheme termed $\textup{D}^{\textup{3}}$T-GAN, addressing the cross-domain GANs transfer in few-shot image generation. Specifically, we design two individual strategies to transfer knowledge between generators and  discriminators, respectively. To transfer knowledge between generators, we conduct a data-dependent transformation, which projects and reconstructs the target samples into the source generator space. Then, we perform knowledge transfer from transformed samples to generated samples. To transfer knowledge between discriminators, we design a multi-level discriminant knowledge distillation from the source discriminator to the target discriminator on both the real and fake samples. Extensive experiments show that our method improve the quality of generated images and achieves the state-of-the-art FID scores on commonly used datasets.

\end{abstract}

	\ccsdesc[500]{Computing methodologies~Reconstruction}
	\ccsdesc[500]{Computing methodologies~Image representations}

\keywords{Few-shot Image Generation, Data-dependent Knowledge Transfer, Projection and Reconstruction}

\maketitle

\section{Introduction}

Recent years have witnessed the development of generative adversarial networks (GANs)~\cite{goodfellow2014generative, karras2019style, karras2020analyzing, brock2018large, radford2015unsupervised, zhang2019self, zhao2016energy, tang2020unified, li2019asymmetric, wang2019u, ho2020sketch}, which learn a function to generate high-dimension images from randomly sampled noises. To be capable of generating realistic images, GANs always rely on a large amount of training samples to fit the real distribution of given categories. However, in real-world scenarios, collecting images is notoriously difficult, due to the privacy protection and expensive labor costs~\cite{fan2021federated, wang2019fewshotvid2vid}. In few-shot scenarios, \ie, training with only dozens or hundreds of samples, the huge imbalance between the limited amount of training data and the high dimension of images makes the learning more difficult. This would disrupt the training and make the model overfitting to the given samples~\cite{zhao2020differentiable, karras2020training}, resulting in poor image quality and low diversity. Thus, it is critical to achieve a well-trained GAN model in few-shot image generation~\cite{phaphuangwittayakul2021fast, guo2020few}.

\begin{figure}[!t]
	\centering
	\includegraphics[width=.75\linewidth]{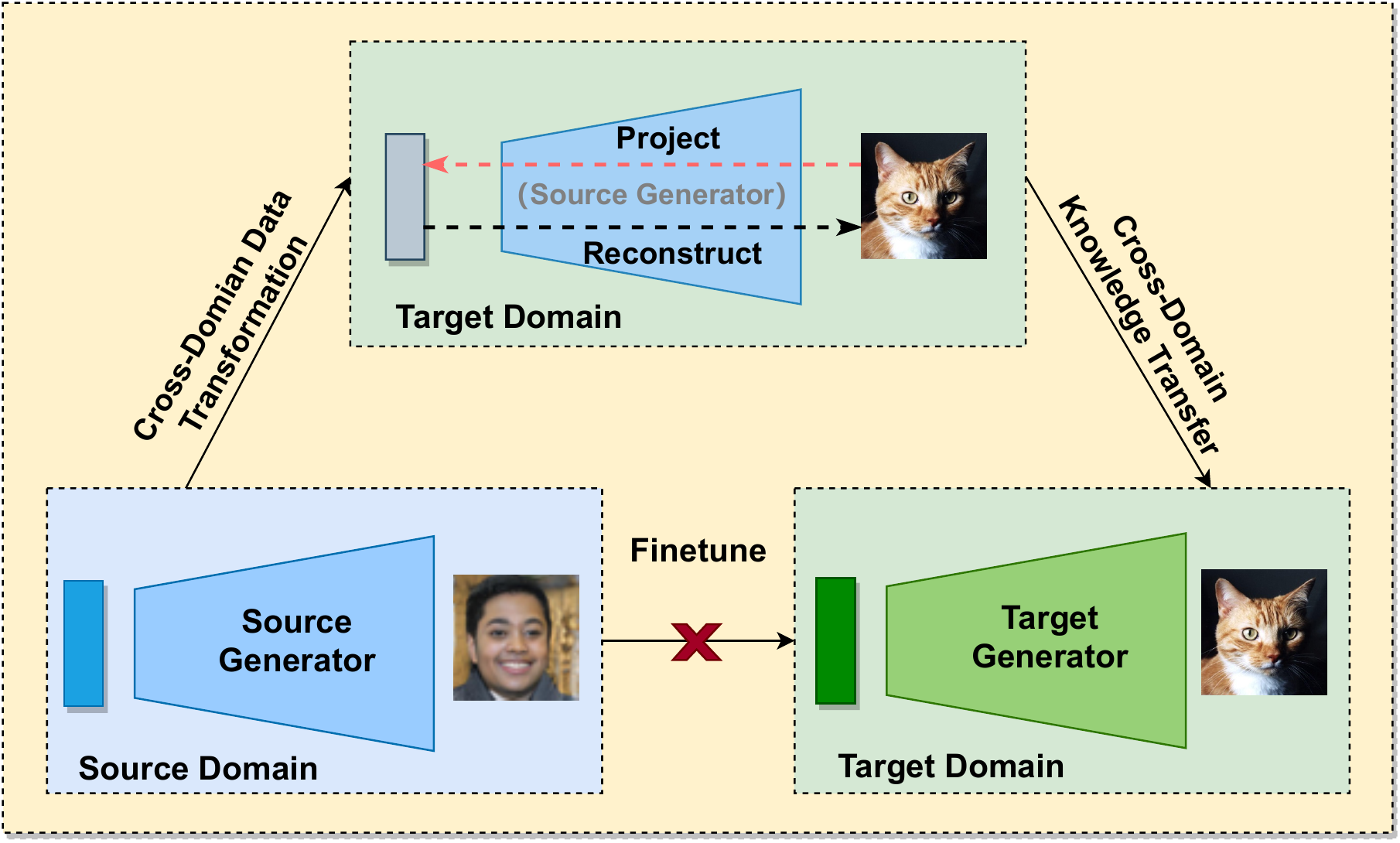}
	\caption{The proposed method $\textup{D}^{\textup{3}}$T-GAN. We perform a novel transfer scheme: cross-domain data-transformation and cross-domain knowledge transfer.}
	\label{fig:motivation}
\end{figure}


In the literature, efforts have been made to empower few-shot learning~\cite{jiang2020few, gao2020pairwise, zhu2021few, wang2019few} in GAN models through transfer learning~\cite{wang2018transferring, noguchi2019image, mo2020freeze, mangla2020data, li2020few, karras2020training, wang2020minegan}. Specifically, a straightforward way is to pre-train a model in the data-rich source domain and directly transfer it to the data-deficient target domain~\cite{wang2018transferring}. Furthermore, several works~\cite{noguchi2019image, wang2020minegan, robb2020few, mo2020freeze} attempt to reduce the model capacity and constrain the optimizable parameters space, which is compatible with few training samples. Since they utilize the prior knowledge of the source domain to enhance the target model learning directly, the image instances of the target domain are not taken into consideration. In this way, these transfer methods are data-independent. Essentially, these approaches can be considered as a warm start for target model training. As a result, when two domains are distant, \ie, different categories, they are prone to suffer from the domain gap. Then, a question arises, how to perform knowledge transfer while not affected by the domain gap? 
In this paper, our goal is to design a data-dependent transfer scheme, realizing cross-domain transfer of generative models.

We address the few-shot generation with a novel knowledge transfer scheme termed Data-Dependent Domain Transfer GAN ($\textup{D}^{\textup{3}}$T-GAN). Overall, we propose two different transfer pipelines for the two modules of a GAN, generator and discriminator. In terms of the generator transfer, we perform a two-stage pipeline: cross-domain data transformation, and cross-domain knowledge transfer. Specifically, we first construct a transformation projecting the target samples into source latent space and reconstruct them into multiple-layer features as the supervision. Then, we impose constraints on the transformed data and original data by transferring the knowledge contained in the reconstructed features to the target model. The transformation is achieved through GAN-Inversion, which has been proved that abundant of latent codes corresponding to other domains can be found in a well-trained generator~\cite{abdal2019image2stylegan}. Thus, the transformed samples are correlated with but different from the original samples. For example, as shown in~\cref{fig:motivation}, the cat (target) images are projected into the human (source) domain and reconstructed into multiple layers of features. These samples achieve the feature diversity, \ie, different representation in other domains, while maintaining the semantic consistency, \ie, reconstruction of the same samples, leading to a more effective transfer effect.

\begin{figure}[t]
	\centering
	\includegraphics[width=.75\linewidth]{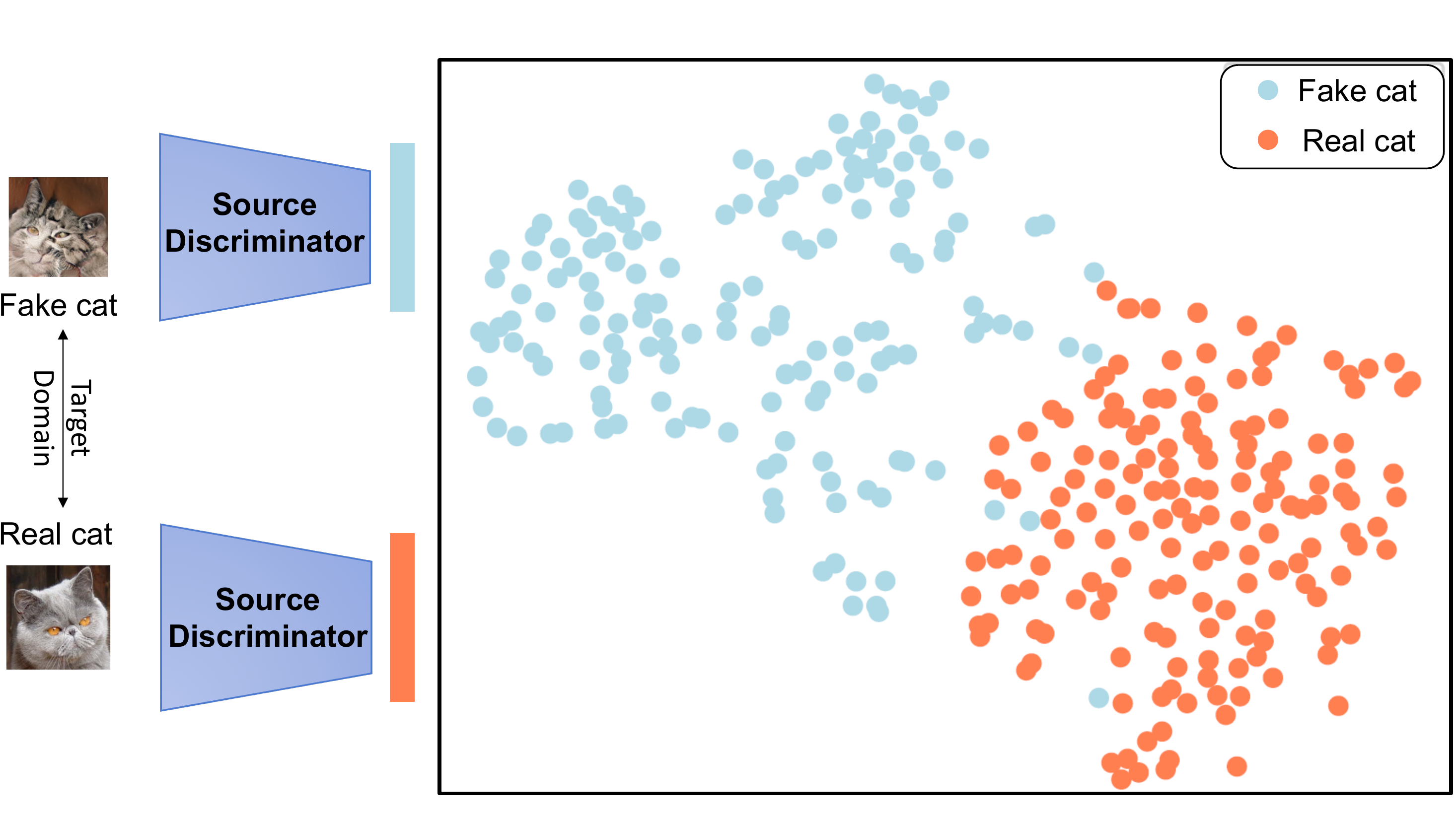}
	\caption{Motivation of our proposed transfer scheme on discriminator. A discriminator trained on source domain can distinguish the Real or Fake of images in target domain. We demonstrate the T-SNE results of the discriminator output features on real and fake cat images. Source discriminator: a well-trained discriminator on FFHQ datasets~\cite{karras2019style}; Real cat: examples on the cat image in Animal Face dataset~\cite{si2011learning}; Fake cat: examples generated by a cat face generator.}
	\label{fig:motivation_second}
\end{figure}

Since the GAN model achieves image generation through the adversarial training between generator and discriminator, the learning of discriminator is also critical. The essence of the discriminator is the distance measure between domains that it aims to distinguish the given image is from real domain or fake domain. Interestingly, we discover that this kind of knowledge, \ie, discriminant ability, is generalized between categories. As shown in~\cref{fig:motivation_second}, a well-trained source discriminator \eg, trained on human faces, is able to distinguish the generated fake cat and real cat images. In this way, we directly inherit the knowledge learned from the multiple layers of source discriminator through knowledge distillation. This is also regarded as a data-dependent domain transfer that we only distill the features depending on both the target samples and generated samples. Therefore, while training with few samples, layers of the target discriminator also learn the knowledge from the source discriminator, achieving a discriminant ability transfer.


In conclusion, the main contributions of this paper are summarized as follows:
\begin{itemize}[leftmargin=*]
    \item We propose $\textup{D}^{\textup{3}}$T-GAN, a novel transfer scheme for GAN models training in few-shot scenarios. We conduct a cross-domain data transformation, and perform a cross-domain knowledge transfer.
	\item We demonstrate the domain generalization of discriminant ability in a well-trained discriminator, and propose a direct transfer inheriting knowledge from the source discriminator to the target one.
    \item Extensive experiments show that our method improves the image quality and achieves the state-of-the-art performance on commonly used datasets.
\end{itemize}

\section{Related Work}

\subsection{Generative Adversarial Networks}
Generative Adversarial Networks (GANs) have been a hot research topic and recent efforts are centered on studying how to train a better GAN for generation and how to apply well-trained GANs to real-world applications. In this subsection, we will introduce several GAN related works in these two aspects as follows.

\noindent\textbf{From-scratch GANs training}.
A GAN model consists of two modules: a generator and a discriminator, both of which achieve a min-max game by iterative adversarial training. However, such a non-convex optimization process is difficult, and always leads to mode collapse and training instability. To alleviate the above problems, several methods focused on optimizing the GAN models through constructing more advanced architectures. For example, ProGAN~\mbox{\cite{karras2017progressive}} is an extension to the standard GAN that allows generation of large high-quality images with a progressive growing pipeline. StyleGAN~\mbox{\cite{karras2019style}} and StyleGANv2~\mbox{\cite{karras2020analyzing}} further improve upon the existing architecture of Generator network with a style mapping to achieve better results. BigGAN~\mbox{\cite{brock2018large}} is able to synthesize natural images according to a given object class. In addition, other works improve the training instability through designing better training strategies in different aspects, which include difference measure between distribution~\mbox{\cite{arjovsky2017wasserstein, nowozin2016f}}, gradient penalties~\mbox{\cite{gulrajani2017improved, mescheder2018training}}, spectral normalization~\mbox{\cite{miyato2018spectral}}, and consistency regularization~\mbox{\cite{zhang2019consistency, zhao2020improved}}. These regularization techniques penalize the changes on the output of discriminators, which improves the quality of the generated images.

\noindent\textbf{Well-trained GANs applications --- GAN Inversion}.
GAN inversion plays an import role in well-trained GAN applications such as real image reconstruction and manipulation. It aims at finding the latent code that is able to recover the input image through an inverted algorithm. In general, there exists two types of approaches. 1) Learning-based methods: learning an extra encoder that maps the images to the latent codes. In general, the encoder is trained to project the input image into the latent code that is able to output the reconstructed image through a fix generator~\mbox{\cite{zhu2016generative, zhu2020domain}}. 
2) Optimization-based methods: updating a randomly initialized latent code in an iterative manner. A straightforward method~\mbox{\cite{abdal2019image2stylegan}} is to optimize the latent code with image reconstruction loss and perceptual loss~\mbox{\cite{dosovitskiy2016generating, gu2020image}} between the reconstructed image and given image. 
Experiments have revealed that the second embedding algorithm is capable to recover samples beyond the original domain. Thus, in this paper, we propose to perform the cross-domain data transformation using the optimization-based inversion technique.

\subsection{Few-shot image generation}
Training a GAN with few samples, also known as few-shot image generation has received widespread attention in recent years. In this subsection, we revisit several definitions of few-shot image generation in recent studies, and divide them into three parts according to the training setting listed as follows.

\noindent\textbf{From a large dataset to a small dataset}.
Several methods proposed to achieve the few-shot image generation through transfer learning. They transferred the prior knowledge leaned in a well-trained model on a large dataset, to the target model on a small dataset. Initially, the most straightforward solution is to transfer knowledge by fine-tuning a pre-trained GAN model~\mbox{\cite{wang2018transferring}}. Moreover, several works attempted to restrict the space of trainable parameters of the generators, including: 1) only updating the normalization layers (\eg, batch normalization)~\mbox{\cite{noguchi2019image}}; 2) fixing the generator and mapping the random noises to the latent codes space through a trainable miner network~\mbox{\cite{wang2020minegan}}; 3) preserving part of the parameters with elastic weights consolidation (EWC) regularization~\mbox{\cite{li2020few}}; 4) only training the highly-expressive parameters that modulate orthogonal features of the pre-trained weight space with singular value decomposition (SVD)~\mbox{\cite{robb2020few}}. Similarly, other works propose to restrict the discriminators for a more effective transfer, including: 1) simply freezing the lower layers of discriminator~\mbox{\cite{mo2020freeze}}; 2) reusing the low-level filters and replacing the high-level layers with a trainable smaller network~\mbox{\cite{zhao2020leveraging}}.

\begin{table}[!t]
	\centering
	\caption{Notations}
	\label{tab:notation}
	\begin{tabu} to 0.55\textwidth {X[c]X[4]}
		\toprule
		$\mathcal{X}$ 	& an image space \\
		$\mathcal{X}_{s} / \mathcal{X}_{t}$ & the source / target dataset \\
		$x^{r} / x^{f}$ & a real / fake sample \\
		$x_t / \hat{x_t}$ & a real target / transformed target sample \\
		
		$\mathcal{W}$ 	& an intermediate style code space \\
		$w$ 			& a style code \\
		$W^{+}$			& an extended style code space\\
		$w^{+}$ 		& an extended style code \\ 
		
		$\mathcal{Z}$ 	& an input noise space \\
		$z$             & an input noise \\
		\midrule
		$G_{s}/G_{t}$	& the source / target generator \\
		$F_{s}/F_{t}$	& the set of intermediate features of $G_{s}$ / $G_{t}$\\
		$D_{s}/D_{t}$	& the source / target discriminator \\
		$E_{s}/E_{t}$	& the set of intermediate features of $D_{s}$ / $D_{t}$\\
		\midrule
		$\mathcal{L}^{adv}_G $	& the adversarial loss function for generator \\
		$\mathcal{L}^{adv}_D $	& the adversarial loss function for discriminator \\
		$\mathcal{L}^{dis}_G$		& the distillation loss for generator	\\
		$\mathcal{L}^{dis}_D$		& the distillation loss for discriminator\\
		$\mathcal{L}^{reg}_{G}$	& the regularization loss for generator	\\
		
		\bottomrule
	\end{tabu}
\end{table}

\noindent\textbf{Only on a small dataset}.
Training a GAN model on a small dataset is difficult to converge and prone to cause mode collapse. Augmentation based generation is a straightforward solution for few-shot image generation that uses data augmentation techniques to expand the size of training samples. However, directly applying the classical data augmentation approach will mislead the generator to learn the distribution of the augmented data~\mbox{\cite{tran2021data}}, resulting in the "leaking" of augmentations to the generated samples~\mbox{\cite{karras2020training}}. Therefore, the differentiable transformations~\mbox{\cite{zhao2020differentiable, zhao2020image}} are proposed on both the real and fake samples so that the gradients are able to propagate through the augmentation back to the generator, and the discriminator is regularized without manipulating the target distribution. Moreover, the data efficiency is further improved through an instance discrimination~\mbox{\cite{yang2021data}} that every image instance is set as an independent category to improve the discriminative ability. In this paper, we combine the differentiable module in ~\mbox{\cite{zhao2020differentiable}} with our method for performance improvement.

\begin{figure}[!t]
	\centering
	\begin{subfigure}{\linewidth}
		\centering
		\includegraphics[width=.75\linewidth]{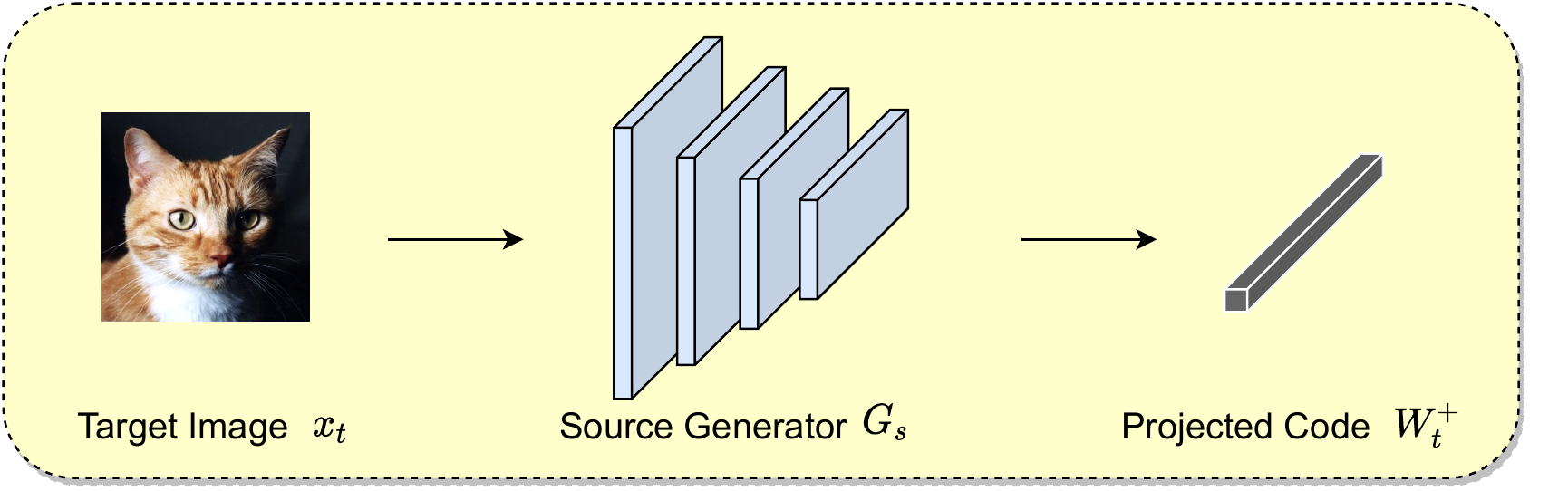}
		\caption{The projection of the target domain images.}
		\label{sfg.NMI1}
	\end{subfigure}
	
	\begin{subfigure}{\linewidth}
		\centering
		\includegraphics[width=.75\linewidth]{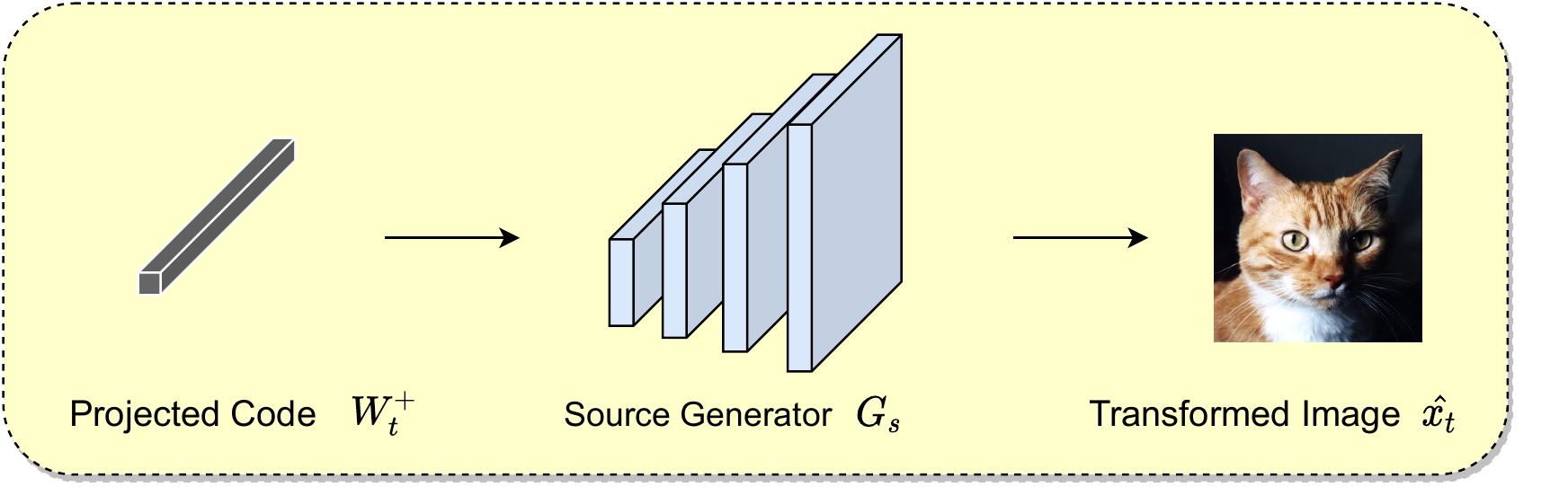}
		\caption{The reconstruction of the projected style codes.}
		\label{sfg.NMI2}
	\end{subfigure}
	\caption{The schematic diagram of the proposed cross-domain data transformation in our method. We project the target domain images into the latent space of source generator and obtain the projected style codes. Then, we reconstruct the projected style code and obtain the transformed images.}
	\label{fig:knowledge_mining}
\end{figure}

\noindent\textbf{From base categories to a novel category}.
Another setting on few-shot image generation is similar as the meta-learning setting~\mbox{\cite{nichol2018first}}. The model is trained on the seen categories with a large amount of labeled data, and evaluated on an unseen category given a few samples. There are several works on this task, including optimization-based \cite{clouatre2019figr, liang2020dawson}, transformation-based \cite{antoniou2017data} and fusion-based methods \cite{bartunov2018few, hong2020matchinggan, hong2020f2gan}. Among them, FIGR \cite{clouatre2019figr} proposed to combine the adversarial learning with optimization-based method Reptile \cite{nichol2018first} to generate new images. DAGAN \cite{antoniou2017data}  input a single image and a random vector, and transformed the input image to get a new image of the same category. F2GAN \cite{hong2020f2gan} input several images and fused them together in the feature space to get a new image of the same category. 
The task settings of these methods are different from ours in two aspects. First, these methods all require an input, \ie, reference image at the inference time. Second, they still require sufficient data for training while we mainly solve how to train a GAN model when given few data. Thus, in this paper, we focus more on the transfer based methods, \ie, transferring models from a large dataset to a small dataset.

\section{Methodology}\label{sec:method}


In this section, we demonstrate the whole proposed method in detail. First, we introduce the problem formulation of this task in~\cref{sec:formulation}. Next, we design two transfer pipelines for knowledge transfer between generators in~\cref{sec:generator_transfer} and knowledge transfer between discriminators in~\cref{sec:discriminator_transfer}. For convenience, \cref{tab:notation} summarizes the notations. The overview of our $\textup{D}^{\textup{3}}$T-GAN is illustrated in~\cref{fig:pipeline}.


\subsection{Problem Formulation}\label{sec:formulation}
Let $X_s$ be the data-rich source domain, and $X_t$ be the data-deficient target domain. These two are different domains of different categories, \ie, $\mathcal{X}_{s} \cap \mathcal{X}_{t} = \emptyset$. In this paper, we only use the training dataset $X_t=\{x^1_t, x^2_t, ..., x^N_t\}$, where $N$ denotes the number of training samples in the target domain, and a well-trained GAN model including a generator $G_s$ and a discriminator $D_s$ in the source domain. We take the StyleGAN as the backbone of the generator network, which includes an input noise space $\mathcal{Z}$, an intermediate style space $\mathcal{W}$ and an output image space $\mathcal{X}$. The generator $G$ maps a random noise $z$ into a style code $w$, and utilizes $w$ to normalize each layer of the backbone to synthesize an image $x$ through AdaIn~\mbox{\cite{huang2017arbitrary}}, where $z\in \mathcal{Z}, w\in \mathcal{W}$, $x\in \mathcal{X}$.

The goal of the few-shot image generation task is to train a GAN model $\{G_t, D_t\}$ that is able to generate realistic images belonging to the target domain using $X_t$. However, training a GAN model from scratch in a low-data regime is difficult to converge and is prone to generate images of poor quality. In this way, we use a transfer-GAN based pipeline, training $\{G_t, D_t\}$ on $X_t$ while transferring the knowledge from $\{G_s, D_s\}$ to $\{G_t, D_t\}$. 

\begin{figure*}[t]
	\centering
	\includegraphics[width=.9\linewidth]{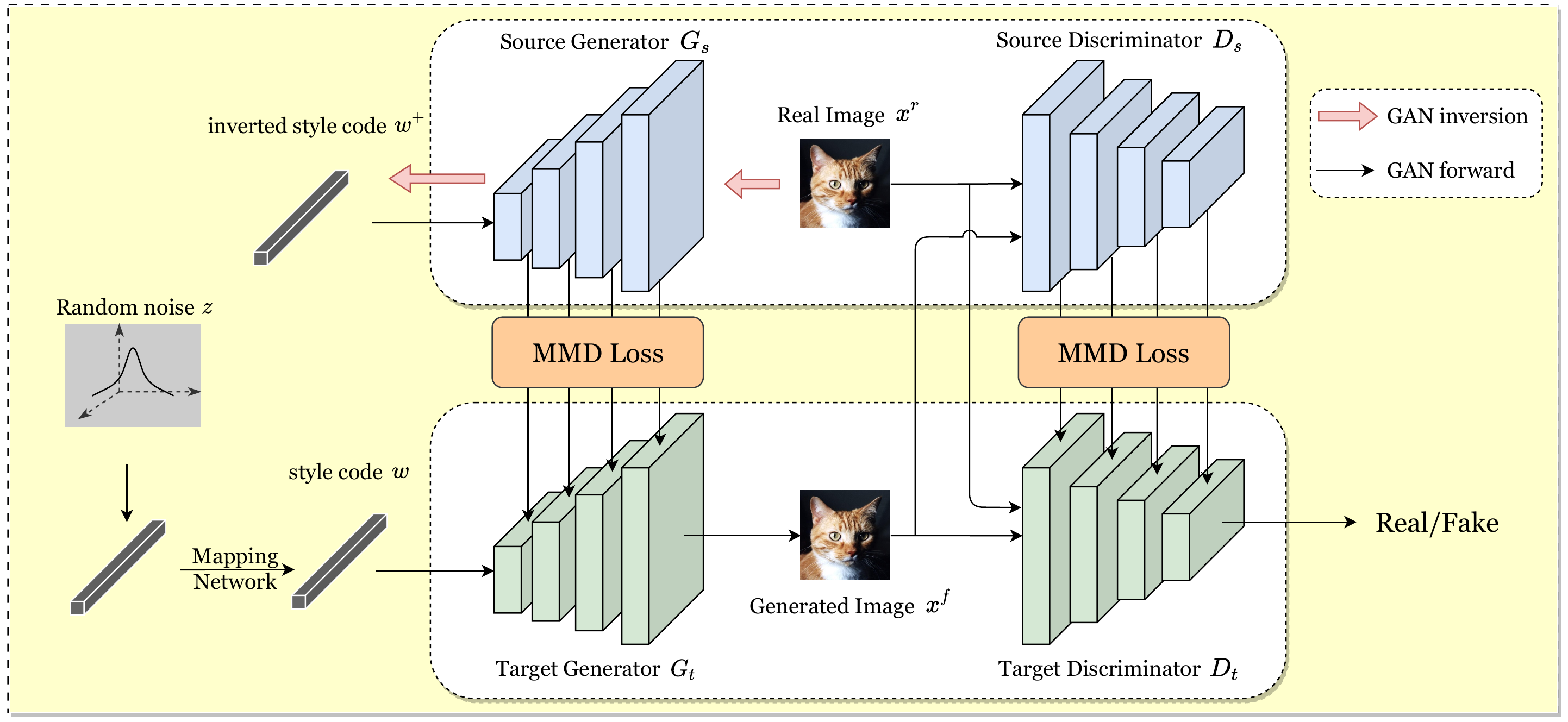}
	\caption{Overview of our proposed $\textup{D}^{\textup{3}}$T-GAN. We select the StyleGAN as the backbone in our method. The networks colored in blue are frozen during the training while those in green are optimized.}
	\label{fig:pipeline}
\end{figure*}
To perform a more effective transfer, we design a data-dependent transfer scheme in a self-supervised manner as shown in \cref{fig:pipeline}. It consists of two pipelines for the generator and discriminator, respectively. In terms of generator transfer, a two-stage transfer pipeline is conducted. First, we transform the target data $x_t$ into $\hat{x_t}$ through GAN-Inversion (See \cref{fig:knowledge_mining}). After obtaining the transformed data $\hat{x_t}$, we align it with the $x_t$ in the feature space of generators $G_s$ and $G_t$. In terms of the discriminator transfer, we perform the feature alignment between source discriminator $D_s$ and target discriminator $D_t$ on the paired input $\{x_t, G_t(z)\}$, where $G_t(z)$ represents the images generated by $G_t$ sampling from Gaussian noise $z$. We keep the consistency of $x_t / G_t(z)$ between the multiple feature spaces in $D_s$ and $D_t$.

\subsection{Generator: Knowledge Transfer via Cross-Domain Data Transformation.}\label{sec:generator_transfer}

In this subsection, we describe the generator transfer in detail, which is a two-stage transfer pipeline in a self-supervised manner. First, we transform the target data from target domain to source domain through GAN-Inversion. Then, we transfer the knowledge of the transformed data to the target domain.

\begin{figure}[!t]
	\centering
	\includegraphics[width=.75\linewidth]{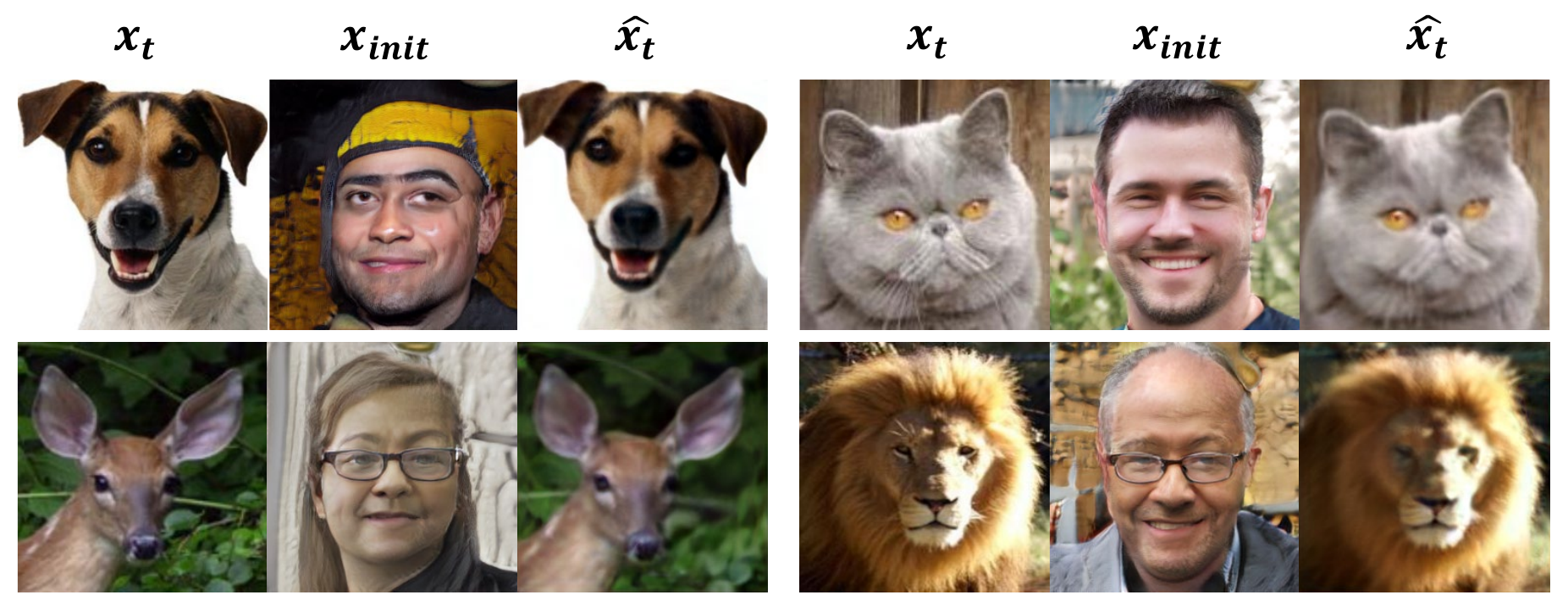}
	\caption{Data transformation results of different categories generated by the source generator trained on FFHQ dataset through GAN-Inversion. $x_t$ refers to the input target image. $x_{init}$ refers to the image corresponding to the initialized random noise of the well-trained generator. $\hat{x_t}$ refers to the final inversion results.}
	\label{fig:inversion}
\end{figure}

\vspace{0.5em}
\noindent\textbf{Cross-domain data transformation}. 
To perform the cross-domain data transformation, we reconstruct the target image in the source generator through projecting and reconstructing, see \mbox{\cref{fig:knowledge_mining}}. Specifically, we embed the target images in the source latent space of generator $G_s$, and search for the latent codes that could synthesize the target images through the source model. The goal of data transformation is to construct data that is correlated with but different from the original data. In order to keep the visual consistency, we follow~\cite{abdal2019image2stylegan} and extend the style code space $W$ of the generator into $W^{+}$, which is a concatenation of 512-dimensional style code vectors. As shown in~\cref{fig:knowledge_mining}, given a target image $x_t\in \mathcal{X}_t$, we embed it into the code space $W^{+}$ by optimizing a random initialized latent code. During the optimization, we adopt a combination of a pixel-wise reconstruction loss and a perceptual loss:
	\begin{equation}\label{con:gan_inversion}
	w^{+}_t = \mathop{\mathrm{argmin}}\limits_{w^{+}\in W^+}({||\hat{x_t} - x_t||^2_2 + \lambda_1 ||C(\hat{x_t})-C(x_t)||^2_2}),
	\end{equation}
	where $\hat{x_t}=G_{s}(w^{+}_t)$ is generated by $G_s$ with the embedded code $w^{+}_t$ of the target sample, $C(\cdot)$ is a VGG-16 network for extracting the features of given images, $\lambda_1$ is the loss weight.
	In this way, the extended style code $w^+_t\in W^{+}$ corresponding to $x_t$, can be found via the source generator $G_{s}$. It is worth noting that, the transformed data $\hat{x_t}$ and the original data $x_t$ are visually and semantic consistent, as shown in~\mbox{\cref{fig:inversion}}. 

\begin{figure*}[!t]
	\centering
	\begin{subfigure}{0.22\linewidth}
		\centering
		\includegraphics[width=\linewidth]{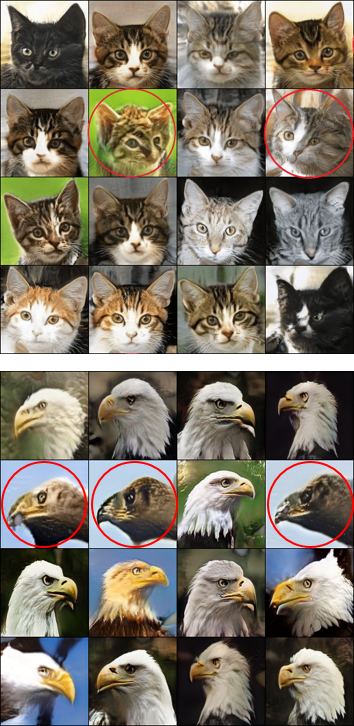}
		\caption{TransferGAN}
	\end{subfigure}
	\hspace{0.3em}
	\begin{subfigure}{0.22\linewidth}
		\centering
		\includegraphics[width=\linewidth]{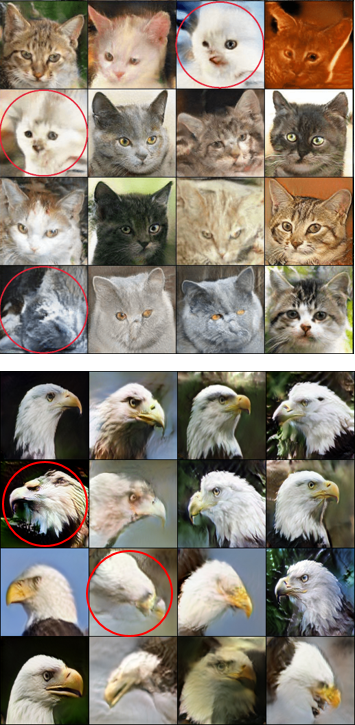}
		\caption{FreezeD}
	\end{subfigure}
	\hspace{0.3em}
	\begin{subfigure}{0.22\linewidth}
		\centering
		\includegraphics[width=\linewidth]{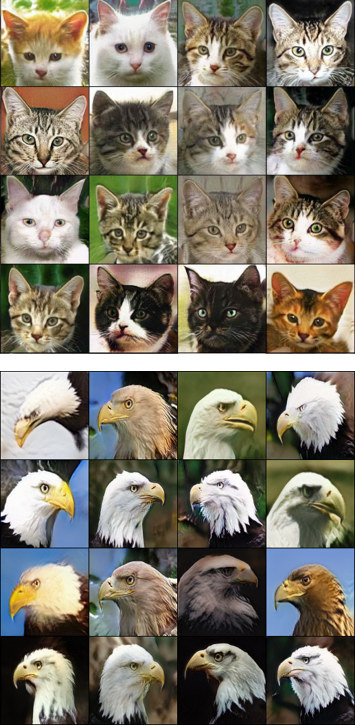}
		\caption{Ours}
	\end{subfigure}%
	\hspace{0.6em}
	\begin{subfigure}{0.22\linewidth}
		\centering
		\includegraphics[width=\linewidth]{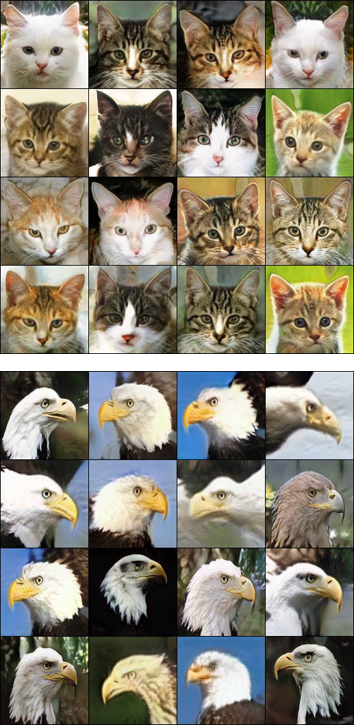}
		\caption{Ours+Diffaugment}
	\end{subfigure}%
	\caption{Qualitative results of our methods compared with TransferGAN~\cite{wang2018transferring} and FreezeD~\cite{mo2020freeze}. This is done by randomly sampling the input noise vectors. The results show that our method is prone to generate more regular cat faces.}
	\label{fig:sample}
\end{figure*}

\vspace{0.5em}
\noindent\textbf{Cross-domain knowledge transfer.}
After obtaining the transformed data $\hat{x_t}$, we propose to align it with $x_t$ that is used to train the target generator $G_t$. To perform knowledge transfer, instead of imposing consistency constraints on two images, we align the feature spaces of two domains. Specifically, with the obtained latent code $w^{+}_t$, we input it into $G_{s}$ and get the intermediate features, 
\begin{equation}
	F_{s}(w^+_t)=\{f^1_{s}(w^+_t), f^2_{s}(w^+_t), ..., f^n_{s}(w^+_t)\},
\end{equation}
where $n$ is the number of layers in source generator $G_s$, $f^i_{s}(w^+)$ is the output of the $i$-th layer. 
When training $G_{t}$, the parameters of $G_{t}$ is initialized to be the same as $G_{s}$. Then, we input a randomly sampled noise $z_t$ to $G_{t}$ and obtain the intermediate features,
\begin{equation}
	F_{t}(z_t)=\{f^1_{t}(z_t), f^2_{t}(z_t), ..., f^n_{t}(z_t)\}.
\end{equation}
Next, we align respective items in $F_{s}$ and $F_{t}$ with Maximum Mean Discrepancy Loss (MMD) as follows, 
\begin{equation}\label{con:mmd_generator}
\begin{aligned}
	\mathcal{L}^{dis}_{G}
	& = \mathcal{L}^{mmd}_G(F_{s}(w^+_t), F_{t}(z_t)) \\
	& = \frac{1}{n}\sum^n_{i=1}||\frac{1}{N}\sum^N_{j=1}\phi(f^{i}_{s}(w^{+,j}_t)) - \frac{1}{N}\sum^N_{j=1}\phi(f^{i}_{t}(z^j_t))||_{\mathcal{H}},
\end{aligned}
\end{equation}
where $\phi(\cdot)$ represents the global average pooling, $w^{+,j}_t$ represents the $j$-th embedded code, $z^j_t$ represents the $j$-th sampled noise, and $\mathcal{H}$ represents the Reproducing Kernel Hilbert Space. 

Moreover, we propose a regularization term for generator training, and the regularization loss $L^{reg}_{G}$ is defined as
\begin{equation}\label{con:reg_generator}
	\mathcal{L}^{reg}_{G} = \mathcal{L}^{mmd}_D(E_{s}(x^r_t), E_{s}(x^f_t)),
\end{equation}
in which, the $\mathcal{L}^{mmd}_D$ is defined as 
\begin{equation}\label{con:mmd_discriminator}
\begin{split}
	& \mathcal{L}^{mmd}_D(E_1(x_t), E_2(x_t)) = \\
	& \frac{1}{m}\sum^m_{i=1}||\frac{1}{N}\sum^N_{j=1}\phi(e^{i}_1(x^j_t)) - \frac{1}{N}\sum^N_{j=1}\phi(e^{i}_2(x^j_t))||_{\mathcal{H}}, 
\end{split}
\end{equation}
where $E(x_t)=\{e^1(x_t), e^2(x_t), ..., e^m(x_t)\}$ refers to the set of features of the target sample in the discriminator, $e^{i}(x^j_t)$ is the feature of $i$-th layer of the $j$-th training sample, $m$ is the number of layers in discriminator, $\phi$ is the global average pooling. For a clearer elaboration, we use $x^r$ to represent the real target image and $x^f$ to represent the generated fake image, \ie, $x^r=x_t, x^f=G_t(z)$, and same below. The purpose of this loss is to make the feature distribution of the generated fake image close to the real image in the source discriminator. 

During the transfer process, adversarial training is necessary for the min-max game. We train $G_{t}$ with $\mathcal{X}_{t}$ with the adversarial loss $L^{adv}_G$, which is defined as: 
\begin{equation}\label{con:adv_generator}
\mathcal{L}^{adv}_G = - \mathbb{E}_{z_t\sim p_z(z_t)}[D_{t}(G_{t}(z_t))],
\end{equation}
where $z_t\sim p_z(z)$ is a sampled noise, $\mathbb{E}$ is the mathematical expectations.

Finally, putting all terms together, the overall optimization of the generator is 
\begin{equation}\label{con:loss_generator}
	\mathcal{L}_{G} = \mathcal{L}^{adv}_G + \lambda_2 \cdot \mathcal{L}^{dis}_G + \lambda_3 \cdot \mathcal{L}^{reg}_G,
\end{equation}
where $\lambda_2$ and $\lambda_3$ are the coefficient weights. 

\begin{table*}[t]
	\centering
	\caption{FID scores under AnimalFace dataset, which consists of totally 20 categories. For each reported method, we train a model on each category and select the snapshot with the best FID score.}
	\setlength{\tabcolsep}{1.0mm}{
		\begin{tabular}{lcccccccccc}
			\toprule
			& Bear  & Cat   & Chicken & Cow   & Deer  & Dog   & Duck  & Eagle & Elephant & Human \\
			\midrule
			TransferGAN~\cite{wang2018transferring} & 58.14 & 51.34 & 111.26 & 101.79 & 48.85 & 92.45 & 138.39 & 92.07 & 89.74 & 118.74 \\
			FreezeD~\cite{mo2020freeze} & 65.65 & 48.09 & 127.66 & 107.21 & 46.47 & 88.73 & 123.37 & 100.49 & 118.95 & \textbf{112.91} \\
			$\textup{D}^{3}$T-GAN (ours)  & \textbf{49.86} & \textbf{45.83} & \textbf{106.58} & \textbf{87.11} & \textbf{45.17} & \textbf{82.02} & \textbf{109.74} & \textbf{76.69} & \textbf{87.49} & 115.95 \\
			\midrule
			& Lion  & Monkey & Mouse & Panda & Pigeon & Pig   & Rabbit & Sheep & Tiger & Wolf \\
			\midrule
			TransferGAN~\cite{wang2018transferring} & 26.05 & 67.24 & 69.86 & 40.91 & 89.01 & 99.18 & 106.43 & 68.79 & 21.58 & 35.44 \\
			FreezeD~\cite{mo2020freeze} & 25.82 & 69.74 & 53.81 & 50.32 & 93.82 & 115.17 & 97.26 & 82.87 & 20.38 & 31.48 \\
			$\textup{D}^{3}$T-GAN (ours)  & \textbf{22.09} & \textbf{55.38} & \textbf{53.28} & \textbf{36.08} & \textbf{85.40} & \textbf{93.41} & \textbf{73.67} & \textbf{62.63} & \textbf{18.56} & \textbf{31.08} \\
			\bottomrule
		\end{tabular}%
	}
	\label{tab:fid_AnimalFace}%
\end{table*}

\subsection{Discriminator: Knowledge Transfer via Multi-Level Features Distillation.}\label{sec:discriminator_transfer}
In this subsection, we introduce the knowledge transfer between discriminators. 
We train $D_{t}$ to distinguish the generated image $x^f=G_t(z_t)$ and target image $x^r=x_t$, while aligning the feature space between $D_{s}$ and $D_{t}$. In vanilla GAN optimization, the discriminator is trained to distinguish $x^f$ and $x^r$ with the adversarial loss $L^{adv}_D$ as,
\begin{equation}
	\label{con:adv_discriminator}
	\mathcal{L}^{adv}_D = \mathbb{E}_{z_t\sim p_z(z_t)}[D_{t}(G_{t}(z_t))] - \mathbb{E}_{p_{data}(x_t)}[D_{t}(x_t)],
\end{equation}
where $p_{data}(x_t)$ is the distribution over $\mathcal{X}_{t}$.

To inherit the knowledge in the well-trained $D_{s}$, we align the projected features of both the real and fake samples in $D_{s}$ and $D_{t}$ by distillation. 
As shown in \mbox{\cref{fig:pipeline}}, we input both the real image $x^{r}$ and fake image $x^{f}$ into both the source discriminator $D_{s}$ and target discriminator $D_{t}$ to obtain: $\{E_{s}(x^r), E_{t}(x^r)\}$, $\{E_{s}(x^f), E_{t}(x^f)\}$. $E_s$ and $E_t$ are the feature sets of $D_s$ and $D_t$, respectively. Then we distill the paired feature sets of fake images and real images at the same time. The distillation loss of the discriminator is presented as,

\begin{equation}\label{con:dis_discriminator}
\begin{split}
	\mathcal{L}^{dis}_{D} = \mathcal{L}^{mmd}_D(E_{s}(x^r), E_{t}(x^r))
	+ \mathcal{L}^{mmd}_D(E_{s}(x^f), E_{t}(x^f)).
\end{split}
\end{equation}
As a result, the optimization of $D_{t}$ is written as,
\begin{equation}\label{con:loss_discriminator}
	\mathcal{L}_{D} = \mathcal{L}^{adv}_D + \lambda_4 \cdot \mathcal{L}^{dis}_D,
\end{equation}
where $\lambda_4$ is a coefficient weight.

\section{Experiments}\label{sec:experiments}

In this section, we demonstrate the effectiveness of our proposed $\textup{D}^{\textup{3}}$T-GAN. First, we introduce the experiment settings in~\cref{sec:experiment setting}. Second, we conduct the experiments on low-shot generation on AnimalFace dataset in subsection~\ref{sec:fid}. Next, we perform qualitative results of our proposed method in~\ref{sec:quality}. Finally, we conduct ablation studies on the transfer strategies in~\ref{sec:ablation studies}. 

\subsection{Experiment setting}\label{sec:experiment setting}

\noindent\textbf{Dataset and Evaluation Metric}.
We conduct our method on the AnimalFace dataset~\cite{si2011learning}, which includes 20 kinds of animal faces such as Cat, Dog, Chicken, Duck, Tiger \textit{etc.}. Each class contains around $100\sim389$ training samples. We set these 20 datasets as different target domains and select the human face dataset Flickr-Faces-HQ Dataset (FFHQ) \cite{karras2019style} as the source domain. It consists of 70,000 high-quality PNG images of human faces. In addition, in the ablation studies, we introduce two categories --- cat and bedroom in the LSUN dataset \cite{yu15lsun} for evaluation. Specifically, LSUN-Bedroom and LSUN-Cat datasets are two large-scale datasets widely used for image generation. We select the whole LSUN-Bedroom dataset (3033042 training samples) for the source StyleGAN training and the part of the LSUN-Cat dataset (5000 training samples) for the number of samples analysis. All the images in the above datasets are preprocessed with a resolution of $256\times 256$.

To evaluate the quality of the generated images, we employ the widely used Fréchet Inception Distance (FID). FID measures the similarity between two given sets, which calculates the difference of the estimated mean and covariance of inception features between the real images and fake images using Inception-V3 \cite{szegedy2016rethinking} network.

\noindent \textbf{Compared Methods}.
Transferring GAN for few-shot generation is a task that has been extensively studied recently. To evaluate the superiority of our method, we compared the proposed $\textup{D}^{\textup{3}}$T-GAN with the following previous methods. TransferGAN \cite{wang2018transferring} directly updates both the generator and discriminator on the target few-shot domain using the well-trained model of the source domain. FreezeD \cite{mo2020freeze} alleviates the overfitting problem by simply freezing parameters in the low-level layer of discriminator on the basis of transferGAN. BSA \cite{noguchi2019image} only updates the batch normalization parameters of the generator on the basis of transferGAN. MineGAN \cite{wang2020minegan} freezes the pre-trained generator and introduces a new miner network to explore the hidden space data distribution.

\noindent \textbf{Implementation Details}.
We utilize StyleGAN and StyleGAN2 as backbones in our method. It is a four-player pipeline including two networks in the source GAN and two in the target GAN. 
The target GAN is initialized with the same weights as the source one. Then we optimize the model with Adam optimizer \cite{kingma2014adam} at a batch size of 16. The learning rate is set to be 0.001 and 0.002 for the generator and for the discriminator, respectively. The exponential decay rates $\beta_1$, $\beta_2$ are set to 0 and 0.99. The hyper-parameters $\lambda_2, \lambda_3, \lambda_4$ are set to 5, 1, 1, respectively. In terms of the GAN-Inversion during the generator transfer, we freeze the well-trained generator and backward it to update the initialized latent code. Specifically, we take advantage of the optimization-based embedding algorithm and set the total iterations to 2000 following~\cite{abdal2019image2stylegan}. The learning rate is initialized to 0.05 and decayed every 500 steps by 10. The loss weights $\lambda_1$ of the perceptual loss in \cref{con:gan_inversion} is set to 5e-5.

\begin{figure*}[!t]
	\centering
	\includegraphics[width=\linewidth]{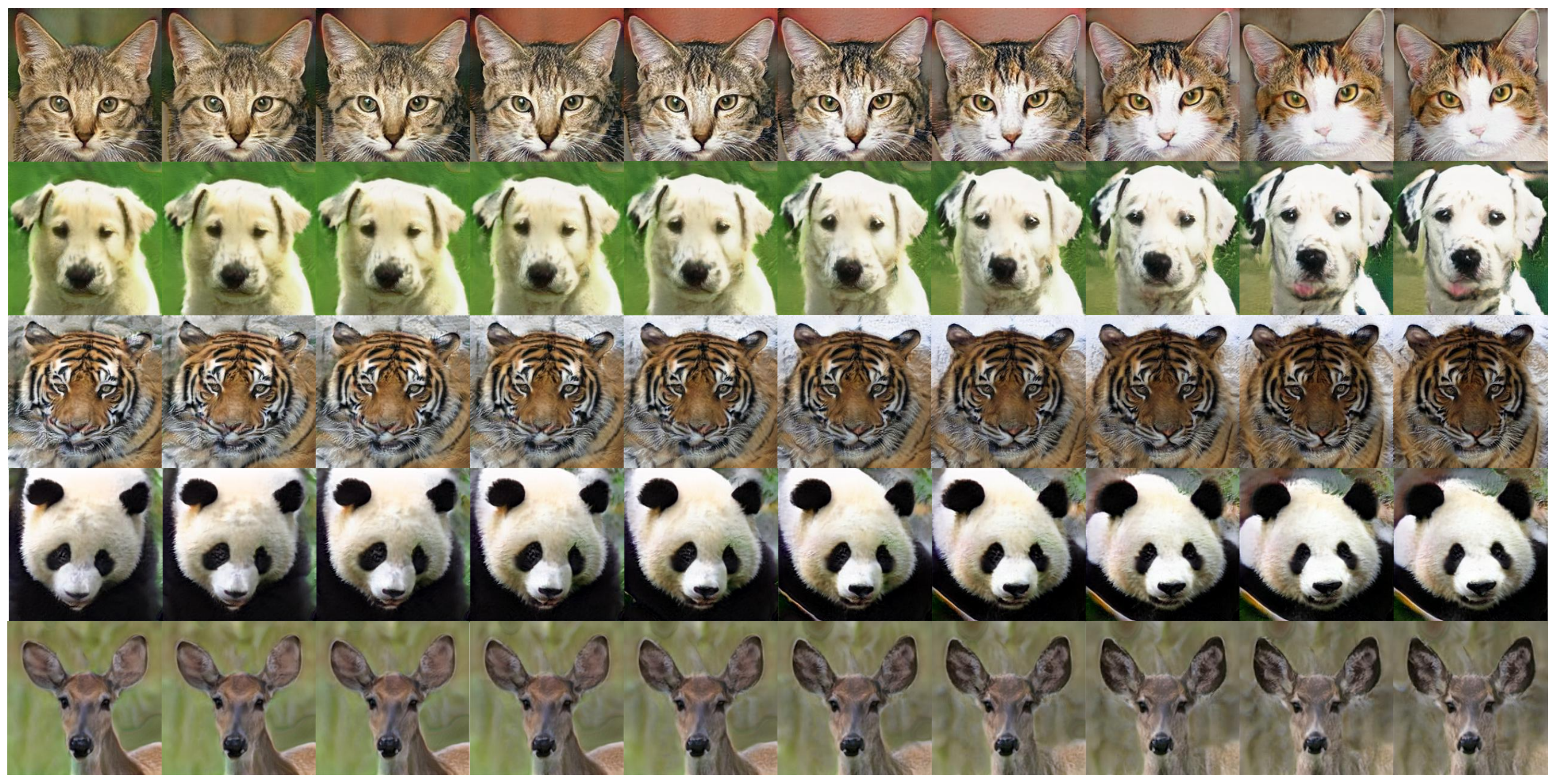}
	\caption{Qualitative results of the style space interpolation in our proposed method. We randomly sample two latent code and generate their intermediate images.}
	\label{fig:interpolation}
\end{figure*}

\subsection{Quantitative Results on Low-shot Datasets}\label{sec:fid}

In this section, we show the quantitative results on low-shot datasets. First, we show the performance of our method on the entire AnimalFace dataset compared with TransferGAN and FreezeD. Next, we compare our method with more related methods on the cat and dog categories based on StyleGAN and StyleGAN2. 

\noindent \textbf{Performance Comparison on Animal Face.}
To evaluate the effectiveness of our method, we conduct experiments on AnimalFace, which includes 20 categories totally. Besides, we also reproduce two previous transfer-gan-based methods TransferGAN~\cite{wang2018transferring} and FreezeD \cite{mo2020freeze} and train them on the entire dataset under the same settings as us. All the source GANs in these methods are well-trained on the FFHQ dataset.

As shown in \cref{tab:fid_AnimalFace}, our method outperforms the previous methods on most of the categories in AnimalFace dataset. The lower FID scores indicate that our method can generate more realistic images. The FID score of our method is about 10 lower than that of FreezeD, and 8 lower than TransferGAN on average, which means a significant improvement of the performance on AnimalFace dataset. Among all categories, Felidae obtains the relatively lower FID scores (\textit{e.g.} cat: $45.83$, tiger: $18.56$, lion: $22.09$) while birds receive the higher scores (\textit{e.g.}, chicken: $106.58$, duck: $109.74$, eagle: $76.69$). This might be the reason that birds and human faces are less similar. Even though, our method still obtains better results than FreezeD on bird categories. For example, we achieve a FID score of $76.69$ on eagle category, with 24 less than that of FreezeD. Besides, we notice that the FID score of the human category ($115.95$) is higher than other categories. Although semantically, the domain distance between the two is relatively close, the result is surprisingly high. This could be the reason that the scale of the face in human category of AnimalFace and FFHQ are different, and the face in the human category is unclear. Specifically, images in AnimalFace-Human contain human head with shoulders and neck while images in FFHQ only include the whole face.

\begin{table}[!t]
	\centering
	\caption{Comparison with State-of-the-art Methods on Cat and Dog datasets. We evaluate all the methods on two backbones. It is obvious that our method outperform others on both the StyleGAN and StyleGAN2.}
	\setlength{\tabcolsep}{2.5mm}{
		\begin{tabular}{l|cc|cc}
		\toprule
		Backbone & \multicolumn{2}{c|}{StyleGAN} & \multicolumn{2}{c}{StyleGAN2} \\
		\midrule
		Dataset & Cat   & Dog   & Cat   & Dog \\
		\midrule
		TransferGAN~\cite{wang2018transferring} & 51.34 & 92.45 & 58.86 & 75.40 \\
		BSA~\cite{noguchi2019image} & 55.02 & 114.81  & ---  & --- \\
		MineGAN~\cite{wang2020minegan} & 53.56 & 95.59 & 48.65 & 77.39  \\
		FreezeD~\cite{mo2020freeze} & 48.09 & 88.73 & 54.94 & 73.17  \\
		$\textup{D}^{\textup{3}}$T-GAN (ours)  & \textbf{45.83} & \textbf{82.02} & \textbf{44.80} & \textbf{70.61} \\
		$\textup{D}^{\textup{3}}$T-GAN (ours)+DiffAugment & \textbf{45.99} & \textbf{76.41} & \textbf{42.07} & \textbf{56.73} \\
		\bottomrule
	\end{tabular}}
	\label{tab:fid_sota}%
\end{table}%

\noindent \textbf{Performance Comparison with Related Methods.}
In this subsection, we compare our method with TransferGAN~\cite{wang2018transferring}, MineGAN~\cite{wang2020minegan}, BSA~\cite{noguchi2019image} and FreezeD~\cite{mo2020freeze}. All the methods are replicated by ourselves under the same settings based on two backbones, StyleGAN and StyleGAN2. As shown in~\cref{tab:fid_sota}, our method achieves the highest performance \ie, the lowest FID scores, on both the cat (45.83 based on StyleGAN and 44.80 based on StyleGAN2) and dog (82.02 based on StyleGAN and 70.61 based on StyleGAN2) categories. Among these previous works, MineGAN tries to introduce a new miner network to explore the hidden space data distribution in the pre-trained network. However, when two domains are distant, this implicit exploration is difficult to find the target distribution. Differently, we explicitly represent the transferred knowledge which is correlated with the target domain through GAN-Inversion, and transfer this knowledge to the target domain. Our method has a significant performance improvement over MineGAN, reducing nearly 8 FID scores and 13 FID scores based on StyleGAN on cat, dog category, respectively. In addition, our method can be also combined with the augmentation based methods for training samples expansion, which achieves a better result with the DiffAugment operation.  

\subsection{Qualitative Results of Low-shot Generation}\label{sec:quality}

In this section, we demonstrate the qualitative results of our method by means of random sampling generation and interpolating generation. 

\begin{figure}[t]
	\centering
	\includegraphics[width=.6\linewidth]{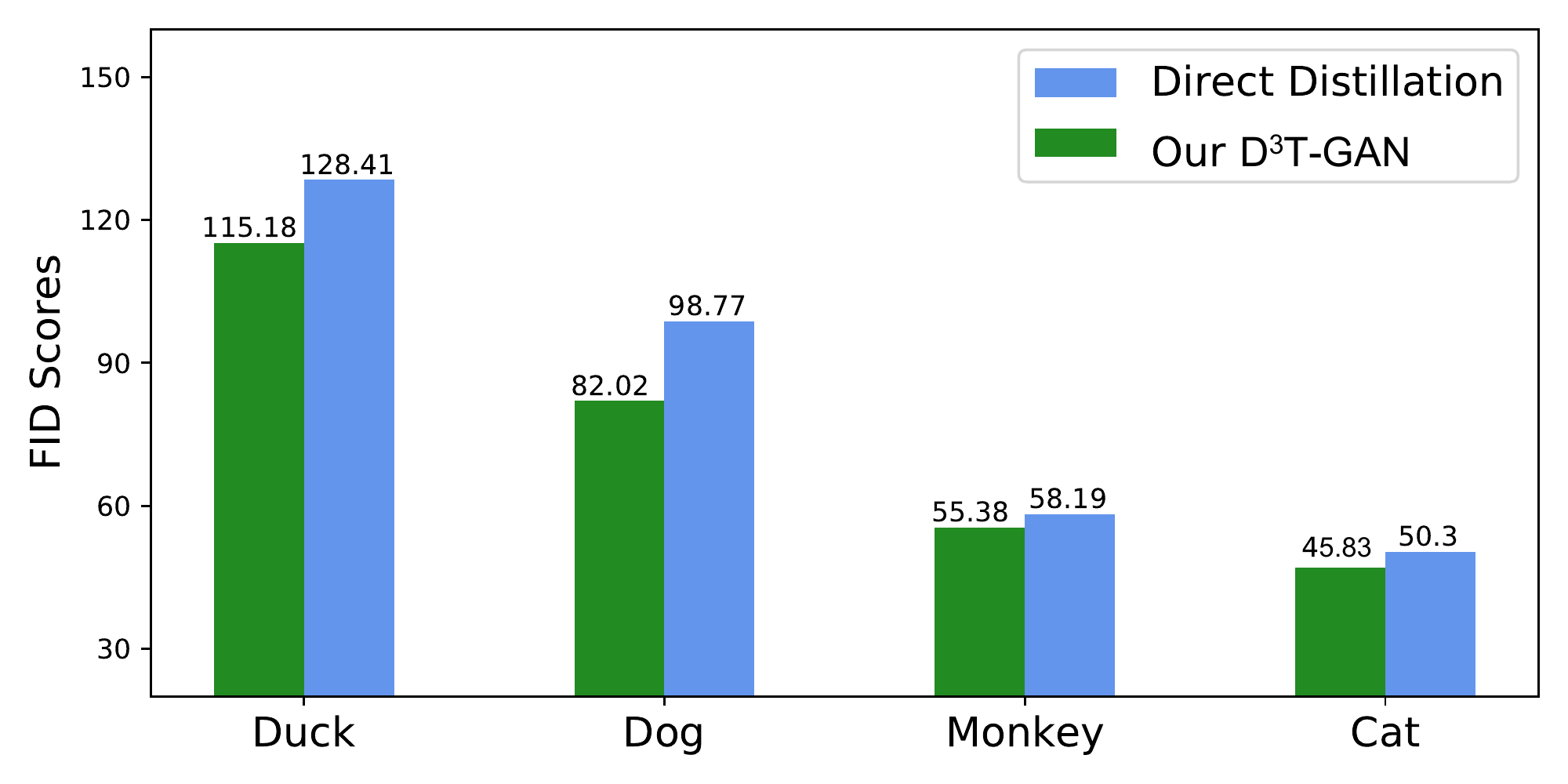}
	\caption{The FID scores of two distillation methods. ''Direct distillation'': directly distilling the features from the source generator to the target generator. Ours: first transforming the data into the source domain and then transfer the data-correlated knowledge only.}
	\label{fig:transfer_compare}
	\vspace{-1em}
\end{figure}

\begin{figure*}[!t]
	\centering
	\includegraphics[width=\linewidth]{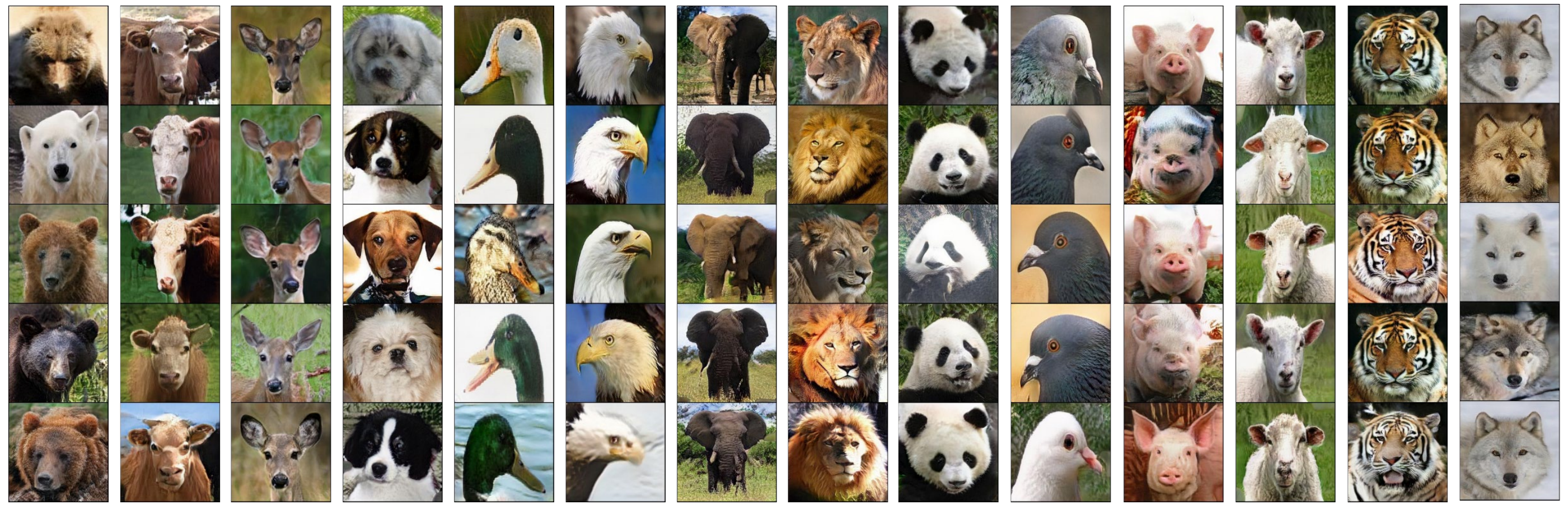}
	\caption{Sampling generation of our method on AnimalFace dataset. We visualize 14 categories in this figure and each column refers to one animal head. }
	\label{fig:sample_small}
\end{figure*}

\begin{figure*}[t]
	\begin{subfigure}{0.325\linewidth}
		\centering
		\includegraphics[width=\linewidth]{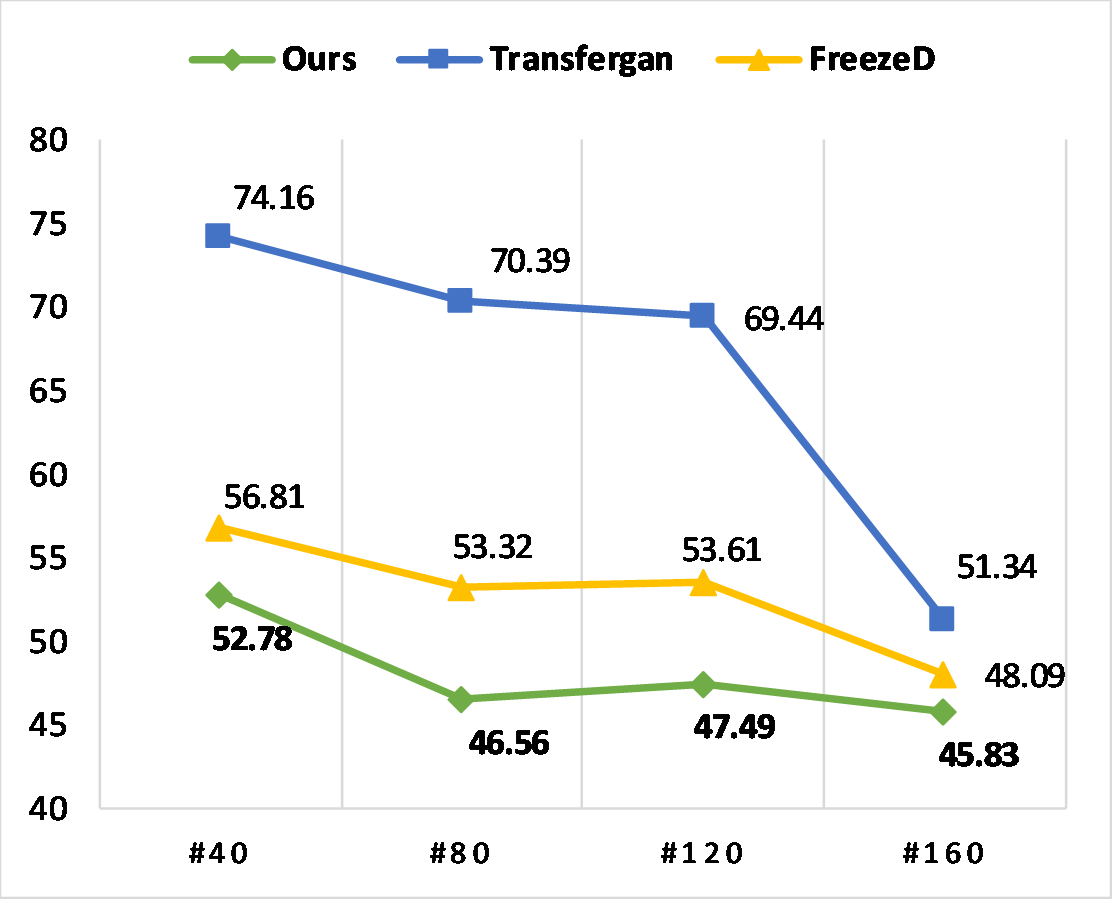}
		\caption{AnimalFace-Cat}
	\end{subfigure}
	\begin{subfigure}{0.32\linewidth}
		\centering
		\includegraphics[width=\linewidth]{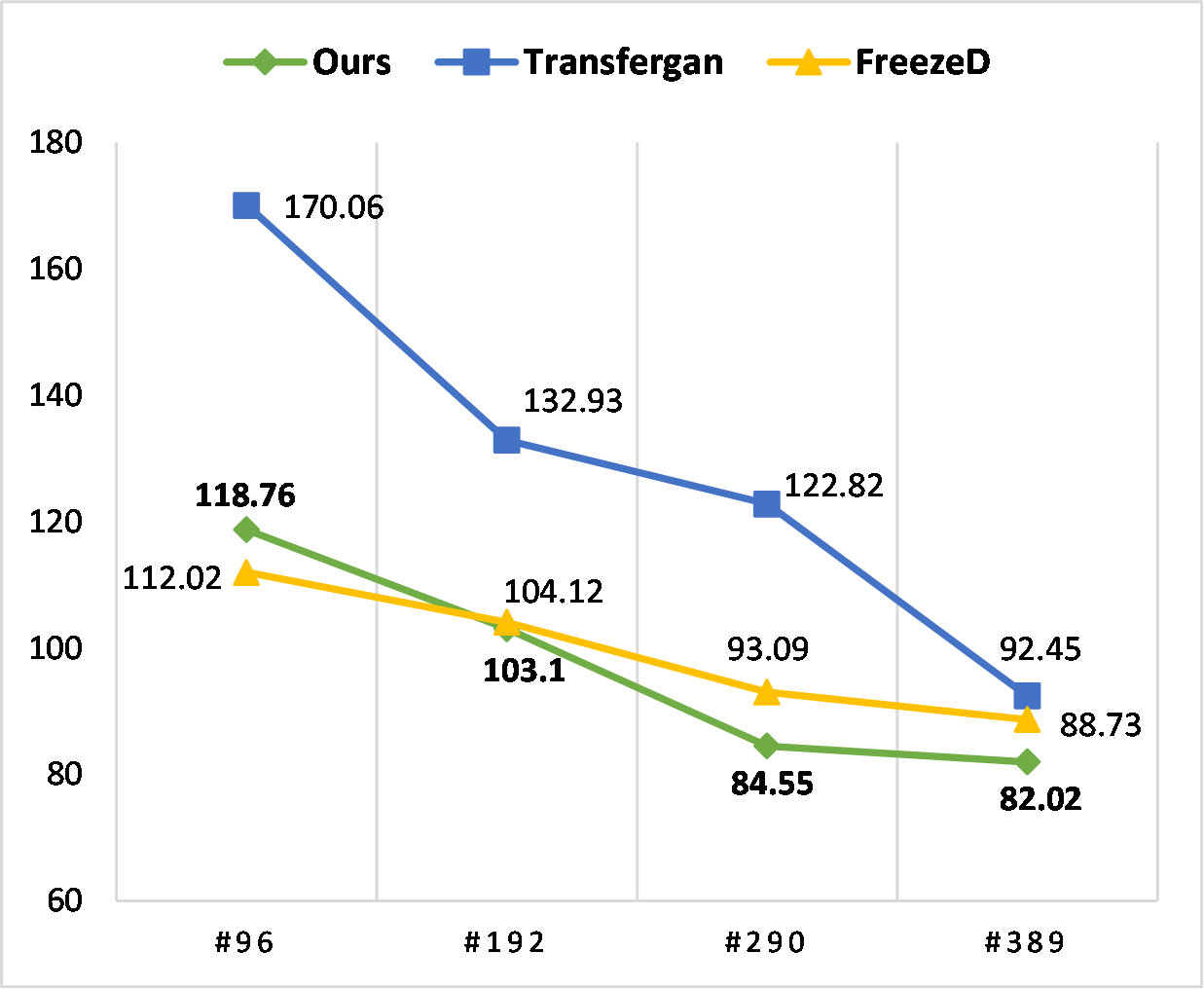}
		\caption{AnimalFace-Dog}
	\end{subfigure}
	\begin{subfigure}{0.32\linewidth}
		\centering
		\includegraphics[width=\linewidth]{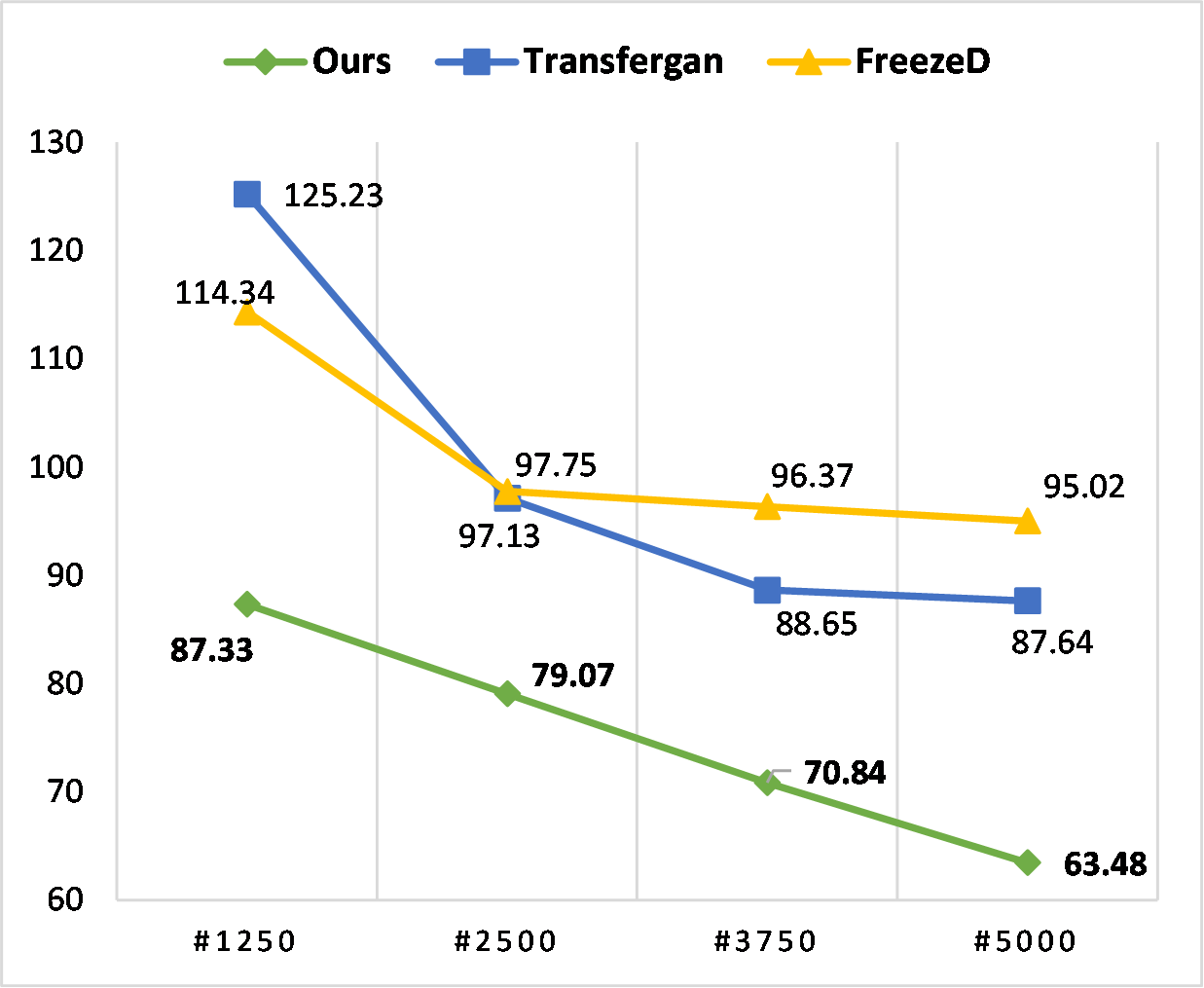}
		\caption{LSUN-Cat}
	\end{subfigure}
	\centering
	\caption{The FID scores of models training with different number of samples. We demonstrate the results of AnimalFace Cat, AnimalFace Dog, and LSUN-Cat. For each category, we select 25\%, 50\%, 75\% and 100\% of the dataset.}
	\label{fig:number_sample}
\end{figure*}

\noindent \textbf{Random Sampling Generation.} 
Through the trained target generator, we randomly sample the input noise vectors under Gaussian distribution and map them to the generated images. As illustrated in \cref{fig:sample}, our method generate cat and eagle images that are more similar with the real images. However, in terms of the TransferGAN and FreezeD, more irregular cat faces are generated than our $\textup{D}^{\textup{3}}$T-GAN. For example, the visualized samples circled in red have strange animal face shapes and even non-face shapes. Our method alleviates this problem since we distill the knowledge of generation process of the source well-trained generator. As shown in \cref{fig:sample_small}, we provide the sampling generation on several categories. Our method is capable of generating realistic images on most of the classes. However, the distribution of the original images in provided datasets highly affects the generated images. \textit{E.g.}, in the tiger category, generated images are similar since the provides samples are close. In the elephant category, generated images are in low quality since most of the original images contain the entire body instead of the head. 

\noindent \textbf{Interpolating Generation.} 
To verify that our method can effectively fit the distribution, we conduct the interpolating generation experiments on five classes, \ie, Cat, Dog, Tiger, Panda and Deer. Specifically, we interpolate two latent codes in the style space $\mathcal{W}$ and show the intermediate results. As shown in~\cref{fig:interpolation}, the interpolation results indicate that our generator generate continuous realistic animal face images, which proves that our method still maintains the smoothness of the manifold in low-shot scenarios.

\subsection{Ablation Studies}\label{sec:ablation studies}

To evaluate the effectiveness of our proposed strategy, we conduct ablation studies on each part our method. 
First, we show the selections of transfer strategies. Next, we study on which layers should be distilled in our method. Then, we demonstrate the impact of the number of training on the transfer results for data-dependent analysis. Finally, we compare whether the domain distance has an impact on the results for cross-domain analysis. 

\vspace{0.5em}
\noindent \textbf{Selection of transfer Strategy}.
Firstly, to evaluate the effectiveness of the proposed self-supervised transfer scheme, we compare ours with a direct transfer on the generator on four categories in AnimalFace dataset. As shown in ~\cref{fig:transfer_compare}, our method performs better than that of ''direct distillation'' on the generator, which directly aligns the two feature sets corresponding to the same sampled noise. Since the animal domains are far apart from the human face domain, directly distilling the knowledge in human face domain will not bring significant performance on the generation on animal domains. As a result, the ''direct distillation'' method receives the higher FID scores especially on duck ($128.41$) and dog ($98.77$) categories. 

\begin{table}[t]
	\centering
	\caption{Ablation Study on the proposed transfer strategies. In this table, $\mathcal{L}^{dis}_D(MMD)$ and $\mathcal{L}^{dis}_G(MMD)$ refer to the proposed distillation loss using MMD as the distribution difference metric. $\mathcal{L}^{dis}_D(L2)$ and $\mathcal{L}^{dis}_G(L2)$ take L2 as the distribution difference metric.}
	\setlength{\tabcolsep}{2.1mm}{
	\begin{tabular}{l|c|c|c|c|c|c}
		\toprule
		Methods & EXP0 & EXP1 & EXP2 & EXP3 & EXP4 & EXP5\\ \midrule
		Pre-train & $\checkmark$ & $\checkmark$ & $\checkmark$ & $\checkmark$ & $\checkmark$ & $\checkmark$\\
		$\mathcal{L}^{dis}_D(MMD)$   &---& $\checkmark$ &---& $\checkmark$ & $\checkmark$ &---\\
		$\mathcal{L}^{dis}_G(MMD)$   &---&---& $\checkmark$ & $\checkmark$ & $\checkmark$ &---\\
		$\mathcal{L}^{dis}_D(L2)$   &---&---&---&---&---&  $\checkmark$ \\
		$\mathcal{L}^{dis}_G(L2)$   &---&---&---&---&---& $\checkmark$ \\
		$\mathcal{L}^{reg}_G$  &---&---&---&---& $\checkmark$ &  $\checkmark$ \\
		\midrule
		Cat   & 51.34 & 49.33 & 49.69 & 48.26 & \textbf{45.83} & 50.68 \\
		Dog   & 92.45 & 87.03 & 87.78 & 86.55 & \textbf{82.02} & 91.71 \\
		\bottomrule
	\end{tabular}%
	}
	\label{tab:ablation_strategy}%
\end{table}%

Moreover, we study the combination of different transfer strategies in our method. As shown in~\cref{tab:ablation_strategy}, $\mathcal{L}^{dis}_D(MMD)$ and $\mathcal{L}^{dis}_G(MMD)$ refer to the proposed knowledge distillation on discriminator and generator using the MMD loss. $\mathcal{L}^{dis}_D(L2)$ and $\mathcal{L}^{dis}_G(L2)$ refer to the proposed knowledge transfer on discriminator and generator using L2 loss. The baseline of our method $\textit{i.e.}$, training the entire GAN model with pre-trained method only achieves the FID scores of $51.34$ on cat and $92.45$ on dog. Adding the three loss items we proposed on the baseline achieves the performance improvements. For example, on dog category, $\mathcal{L}^{dis}_G(MMD)$ and $\mathcal{L}^{dis}_D(MMD)$ both reduce the FID scores of nearly 5. Furthermore, the combination of all the proposed loss items with regularization item achieves the best results, \textit{i.e.}, FID scores of $45.83$ on cat and $82.02$ on dog category. 
In addition, we also make the ablation study on measure of distribution difference, comparing our method using L2 loss. As a result, using MMD performs better in our transfer scheme since it describes the distribution with high-order moments. 

\begin{table}[t]
	\centering
	\caption{FID scores on Cat dataset under different distillation settings. There are 7 layers in both the generator and the discriminator totally. 
		\textbf{Lower Layers} refer to the 1-$st$ layer to the 4-$th$ layer. \textbf{Higher Layers} refer to the 5-$th$ layer to the 7-$th$ layer. \textbf{All Layers} refer to the 1-$st$ layer to the 7-$th$ layer.}
	\setlength{\tabcolsep}{2.2mm}{
		\begin{tabular}{l|l|ccc}
			\toprule
			\multicolumn{2}{c|}{} &\multicolumn{3}{c}{Discriminator} \\ \midrule
			\multicolumn{1}{c|}{} & Distilled Layers & Lower Layers  & Higher Layers & All Layers \\
			\midrule
			\vspace{1mm}
			\multirow{3}{*}{\rotatebox{90}{Generator}}& Lower Layers  & \textbf{45.83}  & 49.31  & --- \\
			\vspace{1mm}
			& Higher Layers & 48.86 & 50.15 & --- \\
			& All Layers & --- & --- & 56.20 \\
			\bottomrule
	\end{tabular}}%
	\label{tab:ablation_layers}%
\end{table}%

\noindent \textbf{Distilled Layers Analysis.}
Secondly, we study on the layers being distilled in the generator and discriminator. Since the generator and discriminator both have 7 layers, we call the first four layers as lower layers and the last three as higher layers. As shown in~\cref{tab:ablation_layers}, we conduct the experiments of different combinations of layers. When we distill the lower layers only, it obtains the highest performance. In contrast, when we distill the higher layers only, it obtains the higher FID scores. Therefore, the low-level semantic information in generator and texture information in discriminator are more proper to be transferred. In addition, if we distill all the layers together, it achieves the worst result in our method's setting. This might be the reason that too many constraints are added to limit the flexibility of networks adjustment.

\noindent \textbf{Number of Training Samples.}
Then, we study how the number of images in target domain will influence the generation quality. We demonstrate the results in \cref{fig:number_sample} on AnimalFace-Cat, AnimalFace-Dog and LSUN-Cat dataset. Notice that the AnimalFace dataset is a low-shot dataset, and thus we add the LSUN-Cat dataset with large number of samples as additional verification. The results suggest that the model performance improves with the increasing number of data. However, our model maintains the performance even with a smaller dataset. For example, when training with half of the AnimalFace-Cat dataset, our model still achieves the FID score of 46.56, which is close to the result training with the entire dataset (45.83). When given more training samples, our method far outperforms the other two, shown the results of LSUN-Cat.

\begin{table}[t]
	\centering
	\caption{Hyper-parameters analysis on $\lambda_2, \lambda_3, \lambda_4$. The \textbf{black bold} results refer to the best FID score and \textcolor[rgb]{0,0,1}{blue} one correspond to the second best FID score.}
	\setlength{\tabcolsep}{3.5mm}{
		\begin{tabular}{c|c|c}
			\toprule
			Hyper-parameters Combination & Cat & Dog  \\
			\midrule
			$\lambda_2=0.1,\quad \lambda_3=1.0,\quad \lambda_4=1.0$  & 47.65 & 92.74 \\
			$\lambda_2=1.0,\quad \lambda_3=1.0,\quad \lambda_4=0.1$  & 50.33 & 86.98 \\
			$\lambda_2=2.0,\quad \lambda_3=1.0,\quad \lambda_4=1.0$  & 47.48 & 84.54 \\
			$\lambda_2=1.0,\quad \lambda_3=1.0,\quad \lambda_4=2.0$  & 47.86 & 91.22 \\
			$\lambda_2=5.0,\quad \lambda_3=1.0,\quad \lambda_4=1.0$ & \textcolor[rgb]{0,0,1}{45.83} & \textcolor[rgb]{0,0,1}{82.02} \\
			$\lambda_2=1.0,\quad \lambda_3=1.0,\quad \lambda_4=5.0$ & 46.95 & 90.22 \\
			$\lambda_2=10,\quad \lambda_3=1.0,\quad \lambda_4=1.0$ & 45.93 & 88.86 \\
			$\lambda_2=1.0,\quad \lambda_3=1.0,\quad \lambda_4=10$ & 48.78 & 89.04 \\
			\midrule
			$\lambda_2=5.0,\quad \lambda_3=0.1,\quad \lambda_4=1.0$ & 46.41 & \textbf{81.95} \\
			$\lambda_2=5.0,\quad \lambda_3=2.0,\quad \lambda_4=1.0$ & \textbf{45.32} & 83.12 \\
			\bottomrule
	\end{tabular}}
	\label{tab:hyper_parameters}%
\end{table}%

\noindent \textbf{Hyper-parameters study.}
Next, we make a simple analysis of the hyper-parameters in $\lambda_2, \lambda_3, \lambda_4$ in \cref{con:loss_generator} and \cref{con:loss_discriminator} to promise the choice of them empirically optimal for the case. We take AnimalFace-Cat and AnimalFace-Dog datasets as example, and evaluate the performance of different combination. The experimental results are shown in \cref{tab:hyper_parameters}. We first fix the regularization weights $\lambda_3=1.0$ and compare different value combinations of distillation weights $\lambda_2$ and $\lambda_4$. The upper half results in \cref{tab:hyper_parameters} indicate that $\{\lambda_2=5.0, \lambda_4=1.0\}$ is the best choice for both cat and dog datasets. It is obvious to find that increasing the weights of generator distillation is more helpful for the performance of increasing that of discriminator. This also shows that our proposed cross-domain knowledge transfer between generators significantly improves GAN transfer. Then, we fix the optimal combination of $\lambda_2, \lambda_4$ and compare the choice of $\lambda_3$. The bottom half results show that different datasets have different adaptations to the regularization weight. Since the optimal result is not much different from that of $\lambda_3=1.0$, we set $\lambda_3$ to 1 in this paper.

\noindent \textbf{Domain Distance Analysis.}
Finally, we make a discussion on how the distance between source and target domain will influence the transfer results. We choose two well-trained StyleGAN models on the large-scale datasets --- LSUN-Bedroom and FFHQ as the source domains, and compare the results of transfer these two to different animal faces. First at all, we measure the FID scores between bedroom (FFHQ) and AnimalFace categories as the domain distance, and report them on the left part of \cref{tab:domain_distance}. Then we fine-tune two pre-trained models on different AnimalFace categories using our method, and report the final FID results on the right part of \cref{tab:domain_distance}. The two-part results illustrate that human face is relatively closer to the animal faces, and intuitively, transfer FFHQ to AnimalFace categories gets better results than transfer LSUN-Bedroom to AnimalFace. In addition, it also reveals such a conclusion. Although bedroom has nothing to do with animal faces in terms of semantics, using it as the pre-trained model still achieves a good transfer effect, which is much better than from scratch training. This could be the reason that images of different domains have similar information in some high-dimensional styles, \eg color. The well-trained generator provides more generalized weights and a short way to generate natural images in such an unstable adversarial training process.

\begin{table}[t]
	\centering
	\caption{Transfer results on domain distance analysis. We demonstrate the transfer results on transfer bedroom and human face to five AnimalFace categories, \ie cat, dog, deer, chicken, and eagle. The left part corresponds to the domain distance between source and target domains measured by FID and the right part is the corresponding results of trained models.}
	\setlength{\tabcolsep}{3.5mm}{
		\begin{tabular}{c|cc|cc}
			\toprule
			& \multicolumn{2}{c|}{Domain Distance} & \multicolumn{2}{c}{Transfer Result} \\
			\midrule
			AnimalFace & Bedroom & FFHQ  & Bedroom & FFHQ  \\
			\midrule
			Cat   & 173.68 & \textbf{166.24} & 55.66 & \textbf{45.83} \\
			Dog   & 139.53 & \textbf{126.34} & 84.22 & \textbf{82.02} \\
			Deer  & 185.64 & \textbf{176.87} & 45.17 & \textbf{45.53} \\
			Chicken & 243.93 & \textbf{233.33} & 108.98 & \textbf{106.58} \\
			Eagle & 187.60 & \textbf{179.18} & 78.78 & \textbf{76.69} \\
			\bottomrule
	\end{tabular}}
	\label{tab:domain_distance}%
\end{table}%

However, we also notice that when comparing vertically in \cref{tab:domain_distance}, the above conclusion does not seem to be valid. For example, the domain distance between FFHQ to AnimalFace-Dog is closer than the distance between FFHQ and AnimalFace-Cat, but the transfer results show that AnimalFace-Cat achieve a better FID score. We guess it could be the reason that when comparing horizontally in the table we calculate the FID scores between different models and same dataset while when comparing vertically, we calculate the FID scores between different models and different datasets. The comparison of the latter seems unfair and does not mean that higher image quality corresponds to lower FID score.

\section{Conclusion}
In this paper, we have designed $\textup{D}^{\textup{3}}$T-GAN, an effective method for transfer generative models between two domains in the few-shot scenarios. We have designed two pipelines for generators and discriminators, respectively. For the generator, we have proposed a novel data-dependent transfer scheme that we first project the target data into the latent space of source generator, and align two feature spaces of source and target generator relying on the transformed data and original data. For discriminator, we directly transfer the knowledge from source to target depending on the given target images and generated images. Experimental results and the extensive visualizations have demonstrated that our method outperforms state-of-the-art methods on both the quality of the generated images and the FID scores. The proposed method is simple to implement while has significant performance gain. 

\bibliographystyle{ACM-Reference-Format}
\bibliography{sample-base}


\begin{thebibliography}{57}


\ifx \showCODEN    \undefined \def \showCODEN     #1{\unskip}     \fi
\ifx \showDOI      \undefined \def \showDOI       #1{#1}\fi
\ifx \showISBNx    \undefined \def \showISBNx     #1{\unskip}     \fi
\ifx \showISBNxiii \undefined \def \showISBNxiii  #1{\unskip}     \fi
\ifx \showISSN     \undefined \def \showISSN      #1{\unskip}     \fi
\ifx \showLCCN     \undefined \def \showLCCN      #1{\unskip}     \fi
\ifx \shownote     \undefined \def \shownote      #1{#1}          \fi
\ifx \showarticletitle \undefined \def \showarticletitle #1{#1}   \fi
\ifx \showURL      \undefined \def \showURL       {\relax}        \fi
\providecommand\bibfield[2]{#2}
\providecommand\bibinfo[2]{#2}
\providecommand\natexlab[1]{#1}
\providecommand\showeprint[2][]{arXiv:#2}

\bibitem[\protect\citeauthoryear{Abdal, Qin, and Wonka}{Abdal
  et~al\mbox{.}}{2019}]%
        {abdal2019image2stylegan}
\bibfield{author}{\bibinfo{person}{Rameen Abdal}, \bibinfo{person}{Yipeng Qin},
  {and} \bibinfo{person}{Peter Wonka}.} \bibinfo{year}{2019}\natexlab{}.
\newblock \showarticletitle{Image2stylegan: How to embed images into the
  stylegan latent space?}. In \bibinfo{booktitle}{\emph{Proc. ICCV}}.
  \bibinfo{pages}{4432--4441}.
\newblock


\bibitem[\protect\citeauthoryear{Antoniou, Storkey, and Edwards}{Antoniou
  et~al\mbox{.}}{2017}]%
        {antoniou2017data}
\bibfield{author}{\bibinfo{person}{Antreas Antoniou}, \bibinfo{person}{Amos
  Storkey}, {and} \bibinfo{person}{Harrison Edwards}.}
  \bibinfo{year}{2017}\natexlab{}.
\newblock \showarticletitle{Data augmentation generative adversarial networks}.
  In \bibinfo{booktitle}{\emph{Proc. ICLR}}.
\newblock


\bibitem[\protect\citeauthoryear{Arjovsky, Chintala, and Bottou}{Arjovsky
  et~al\mbox{.}}{2017}]%
        {arjovsky2017wasserstein}
\bibfield{author}{\bibinfo{person}{Martin Arjovsky}, \bibinfo{person}{Soumith
  Chintala}, {and} \bibinfo{person}{L{\'e}on Bottou}.}
  \bibinfo{year}{2017}\natexlab{}.
\newblock \showarticletitle{Wasserstein generative adversarial networks}. In
  \bibinfo{booktitle}{\emph{Proc. ICML}}. PMLR, \bibinfo{pages}{214--223}.
\newblock


\bibitem[\protect\citeauthoryear{Bartunov and Vetrov}{Bartunov and
  Vetrov}{2018}]%
        {bartunov2018few}
\bibfield{author}{\bibinfo{person}{Sergey Bartunov} {and}
  \bibinfo{person}{Dmitry Vetrov}.} \bibinfo{year}{2018}\natexlab{}.
\newblock \showarticletitle{Few-shot generative modelling with generative
  matching networks}. In \bibinfo{booktitle}{\emph{Proc. AISTATS}}. PMLR,
  \bibinfo{pages}{670--678}.
\newblock


\bibitem[\protect\citeauthoryear{Brock, Donahue, and Simonyan}{Brock
  et~al\mbox{.}}{2019}]%
        {brock2018large}
\bibfield{author}{\bibinfo{person}{Andrew Brock}, \bibinfo{person}{Jeff
  Donahue}, {and} \bibinfo{person}{Karen Simonyan}.}
  \bibinfo{year}{2019}\natexlab{}.
\newblock \showarticletitle{Large scale GAN training for high fidelity natural
  image synthesis}. In \bibinfo{booktitle}{\emph{Proc. ICLR}}.
\newblock


\bibitem[\protect\citeauthoryear{Clou{\^a}tre and Demers}{Clou{\^a}tre and
  Demers}{2019}]%
        {clouatre2019figr}
\bibfield{author}{\bibinfo{person}{Louis Clou{\^a}tre} {and}
  \bibinfo{person}{Marc Demers}.} \bibinfo{year}{2019}\natexlab{}.
\newblock \showarticletitle{FIGR: Few-shot image generation with reptile}. In
  \bibinfo{booktitle}{\emph{arXiv preprint arXiv:1901.02199}}.
\newblock


\bibitem[\protect\citeauthoryear{Dosovitskiy and Brox}{Dosovitskiy and
  Brox}{2016}]%
        {dosovitskiy2016generating}
\bibfield{author}{\bibinfo{person}{Alexey Dosovitskiy} {and}
  \bibinfo{person}{Thomas Brox}.} \bibinfo{year}{2016}\natexlab{}.
\newblock \showarticletitle{Generating images with perceptual similarity
  metrics based on deep networks}. In \bibinfo{booktitle}{\emph{Proc.
  NeurIPS}}, Vol.~\bibinfo{volume}{29}. \bibinfo{pages}{658--666}.
\newblock


\bibitem[\protect\citeauthoryear{Fan and Huang}{Fan and Huang}{2021}]%
        {fan2021federated}
\bibfield{author}{\bibinfo{person}{Chenyou Fan} {and} \bibinfo{person}{Jianwei
  Huang}.} \bibinfo{year}{2021}\natexlab{}.
\newblock \showarticletitle{Federated Few-Shot Learning with Adversarial
  Learning}. In \bibinfo{booktitle}{\emph{arXiv preprint arXiv:2104.00365}}.
\newblock


\bibitem[\protect\citeauthoryear{Gao, Guo, Guan, Liu, Ren, and Chen}{Gao
  et~al\mbox{.}}{2020}]%
        {gao2020pairwise}
\bibfield{author}{\bibinfo{person}{Zan Gao}, \bibinfo{person}{Leming Guo},
  \bibinfo{person}{Weili Guan}, \bibinfo{person}{An-An Liu},
  \bibinfo{person}{Tongwei Ren}, {and} \bibinfo{person}{Shengyong Chen}.}
  \bibinfo{year}{2020}\natexlab{}.
\newblock \showarticletitle{A pairwise attentive adversarial spatiotemporal
  network for cross-domain few-shot action recognition-R2}. In
  \bibinfo{booktitle}{\emph{{IEEE} Trans. Image Process.}},
  Vol.~\bibinfo{volume}{30}. \bibinfo{publisher}{IEEE},
  \bibinfo{pages}{767--782}.
\newblock


\bibitem[\protect\citeauthoryear{Goodfellow, Pouget-Abadie, Mirza, Xu,
  Warde-Farley, Ozair, Courville, and Bengio}{Goodfellow et~al\mbox{.}}{2014}]%
        {goodfellow2014generative}
\bibfield{author}{\bibinfo{person}{Ian~J Goodfellow}, \bibinfo{person}{Jean
  Pouget-Abadie}, \bibinfo{person}{Mehdi Mirza}, \bibinfo{person}{Bing Xu},
  \bibinfo{person}{David Warde-Farley}, \bibinfo{person}{Sherjil Ozair},
  \bibinfo{person}{Aaron Courville}, {and} \bibinfo{person}{Yoshua Bengio}.}
  \bibinfo{year}{2014}\natexlab{}.
\newblock \showarticletitle{Generative adversarial networks}. In
  \bibinfo{booktitle}{\emph{Proc. NeurIPS}}.
\newblock


\bibitem[\protect\citeauthoryear{Gu, Shen, and Zhou}{Gu et~al\mbox{.}}{2020}]%
        {gu2020image}
\bibfield{author}{\bibinfo{person}{Jinjin Gu}, \bibinfo{person}{Yujun Shen},
  {and} \bibinfo{person}{Bolei Zhou}.} \bibinfo{year}{2020}\natexlab{}.
\newblock \showarticletitle{Image processing using multi-code gan prior}. In
  \bibinfo{booktitle}{\emph{Proc. CVPR}}. \bibinfo{pages}{3012--3021}.
\newblock


\bibitem[\protect\citeauthoryear{Gulrajani, Ahmed, Arjovsky, Dumoulin, and
  Courville}{Gulrajani et~al\mbox{.}}{2017}]%
        {gulrajani2017improved}
\bibfield{author}{\bibinfo{person}{Ishaan Gulrajani}, \bibinfo{person}{Faruk
  Ahmed}, \bibinfo{person}{Martin Arjovsky}, \bibinfo{person}{Vincent
  Dumoulin}, {and} \bibinfo{person}{Aaron Courville}.}
  \bibinfo{year}{2017}\natexlab{}.
\newblock \showarticletitle{Improved training of wasserstein gans}. In
  \bibinfo{booktitle}{\emph{Proc. NeurIPS}}.
\newblock


\bibitem[\protect\citeauthoryear{Guo, Zhang, Jiang, Niu, Gu, Zheng, Wang, and
  Zheng}{Guo et~al\mbox{.}}{2020}]%
        {guo2020few}
\bibfield{author}{\bibinfo{person}{Zonghui Guo}, \bibinfo{person}{Liqiang
  Zhang}, \bibinfo{person}{Yufeng Jiang}, \bibinfo{person}{Wenjie Niu},
  \bibinfo{person}{Zhaorui Gu}, \bibinfo{person}{Haiyong Zheng},
  \bibinfo{person}{Guoyu Wang}, {and} \bibinfo{person}{Bing Zheng}.}
  \bibinfo{year}{2020}\natexlab{}.
\newblock \showarticletitle{Few-shot Fish Image Generation and Classification}.
  In \bibinfo{booktitle}{\emph{Global Oceans}}. IEEE, \bibinfo{pages}{1--6}.
\newblock


\bibitem[\protect\citeauthoryear{Ho, Virtusio, Chen, Hsu, and Hua}{Ho
  et~al\mbox{.}}{2020}]%
        {ho2020sketch}
\bibfield{author}{\bibinfo{person}{Trang-Thi Ho}, \bibinfo{person}{John~Jethro
  Virtusio}, \bibinfo{person}{Yung-Yao Chen}, \bibinfo{person}{Chih-Ming Hsu},
  {and} \bibinfo{person}{Kai-Lung Hua}.} \bibinfo{year}{2020}\natexlab{}.
\newblock \showarticletitle{Sketch-guided deep portrait generation}. In
  \bibinfo{booktitle}{\emph{ACM Trans. Multimedia Comput. Commun. Appl.}},
  Vol.~\bibinfo{volume}{16}. \bibinfo{publisher}{ACM New York, NY, USA},
  \bibinfo{pages}{1--18}.
\newblock


\bibitem[\protect\citeauthoryear{Hong, Niu, Zhang, and Zhang}{Hong
  et~al\mbox{.}}{2020a}]%
        {hong2020matchinggan}
\bibfield{author}{\bibinfo{person}{Yan Hong}, \bibinfo{person}{Li Niu},
  \bibinfo{person}{Jianfu Zhang}, {and} \bibinfo{person}{Liqing Zhang}.}
  \bibinfo{year}{2020}\natexlab{a}.
\newblock \showarticletitle{MatchingGAN: Matching-based few-shot image
  generation}. In \bibinfo{booktitle}{\emph{Proc. IEEE ICME}}. IEEE,
  \bibinfo{pages}{1--6}.
\newblock


\bibitem[\protect\citeauthoryear{Hong, Niu, Zhang, Zhao, Fu, and Zhang}{Hong
  et~al\mbox{.}}{2020b}]%
        {hong2020f2gan}
\bibfield{author}{\bibinfo{person}{Yan Hong}, \bibinfo{person}{Li Niu},
  \bibinfo{person}{Jianfu Zhang}, \bibinfo{person}{Weijie Zhao},
  \bibinfo{person}{Chen Fu}, {and} \bibinfo{person}{Liqing Zhang}.}
  \bibinfo{year}{2020}\natexlab{b}.
\newblock \showarticletitle{F2GAN: Fusing-and-Filling GAN for Few-shot Image
  Generation}. In \bibinfo{booktitle}{\emph{Proc. ACM MM}}.
  \bibinfo{pages}{2535--2543}.
\newblock


\bibitem[\protect\citeauthoryear{Huang and Belongie}{Huang and
  Belongie}{2017}]%
        {huang2017arbitrary}
\bibfield{author}{\bibinfo{person}{Xun Huang} {and} \bibinfo{person}{Serge
  Belongie}.} \bibinfo{year}{2017}\natexlab{}.
\newblock \showarticletitle{Arbitrary style transfer in real-time with adaptive
  instance normalization}. In \bibinfo{booktitle}{\emph{Proc. ICCV}}.
  \bibinfo{pages}{1501--1510}.
\newblock


\bibitem[\protect\citeauthoryear{Jiang, Min, Lyu, and Liu}{Jiang
  et~al\mbox{.}}{2020}]%
        {jiang2020few}
\bibfield{author}{\bibinfo{person}{Shuqiang Jiang}, \bibinfo{person}{Weiqing
  Min}, \bibinfo{person}{Yongqiang Lyu}, {and} \bibinfo{person}{Linhu Liu}.}
  \bibinfo{year}{2020}\natexlab{}.
\newblock \showarticletitle{Few-shot food recognition via multi-view
  representation learning}. In \bibinfo{booktitle}{\emph{ACM Trans. Multimedia
  Comput. Commun. Appl.}}, Vol.~\bibinfo{volume}{16}. \bibinfo{publisher}{ACM
  New York, NY, USA}, \bibinfo{pages}{1--20}.
\newblock


\bibitem[\protect\citeauthoryear{Karras, Aila, Laine, and Lehtinen}{Karras
  et~al\mbox{.}}{2018}]%
        {karras2017progressive}
\bibfield{author}{\bibinfo{person}{Tero Karras}, \bibinfo{person}{Timo Aila},
  \bibinfo{person}{Samuli Laine}, {and} \bibinfo{person}{Jaakko Lehtinen}.}
  \bibinfo{year}{2018}\natexlab{}.
\newblock \showarticletitle{Progressive growing of gans for improved quality,
  stability, and variation}. In \bibinfo{booktitle}{\emph{Proc. ICLR}}.
\newblock


\bibitem[\protect\citeauthoryear{Karras, Aittala, Hellsten, Laine, Lehtinen,
  and Aila}{Karras et~al\mbox{.}}{2020a}]%
        {karras2020training}
\bibfield{author}{\bibinfo{person}{Tero Karras}, \bibinfo{person}{Miika
  Aittala}, \bibinfo{person}{Janne Hellsten}, \bibinfo{person}{Samuli Laine},
  \bibinfo{person}{Jaakko Lehtinen}, {and} \bibinfo{person}{Timo Aila}.}
  \bibinfo{year}{2020}\natexlab{a}.
\newblock \showarticletitle{Training generative adversarial networks with
  limited data}. In \bibinfo{booktitle}{\emph{Proc. NeurIPS}}.
\newblock


\bibitem[\protect\citeauthoryear{Karras, Laine, and Aila}{Karras
  et~al\mbox{.}}{2019}]%
        {karras2019style}
\bibfield{author}{\bibinfo{person}{Tero Karras}, \bibinfo{person}{Samuli
  Laine}, {and} \bibinfo{person}{Timo Aila}.} \bibinfo{year}{2019}\natexlab{}.
\newblock \showarticletitle{A style-based generator architecture for generative
  adversarial networks}. In \bibinfo{booktitle}{\emph{Proc. CVPR}}.
  \bibinfo{pages}{4401--4410}.
\newblock


\bibitem[\protect\citeauthoryear{Karras, Laine, Aittala, Hellsten, Lehtinen,
  and Aila}{Karras et~al\mbox{.}}{2020b}]%
        {karras2020analyzing}
\bibfield{author}{\bibinfo{person}{Tero Karras}, \bibinfo{person}{Samuli
  Laine}, \bibinfo{person}{Miika Aittala}, \bibinfo{person}{Janne Hellsten},
  \bibinfo{person}{Jaakko Lehtinen}, {and} \bibinfo{person}{Timo Aila}.}
  \bibinfo{year}{2020}\natexlab{b}.
\newblock \showarticletitle{Analyzing and improving the image quality of
  stylegan}. In \bibinfo{booktitle}{\emph{Proc. CVPR}}.
  \bibinfo{pages}{8110--8119}.
\newblock


\bibitem[\protect\citeauthoryear{Kingma and Ba}{Kingma and Ba}{2015}]%
        {kingma2014adam}
\bibfield{author}{\bibinfo{person}{Diederik~P Kingma} {and}
  \bibinfo{person}{Jimmy Ba}.} \bibinfo{year}{2015}\natexlab{}.
\newblock \showarticletitle{Adam: A method for stochastic optimization}. In
  \bibinfo{booktitle}{\emph{Proc. ICLR}}.
\newblock


\bibitem[\protect\citeauthoryear{Li, Tang, Zhang, Zhang, Li, and Yan}{Li
  et~al\mbox{.}}{2019}]%
        {li2019asymmetric}
\bibfield{author}{\bibinfo{person}{Yu Li}, \bibinfo{person}{Sheng Tang},
  \bibinfo{person}{Rui Zhang}, \bibinfo{person}{Yongdong Zhang},
  \bibinfo{person}{Jintao Li}, {and} \bibinfo{person}{Shuicheng Yan}.}
  \bibinfo{year}{2019}\natexlab{}.
\newblock \showarticletitle{Asymmetric GAN for unpaired image-to-image
  translation}. In \bibinfo{booktitle}{\emph{{IEEE} Trans. Image Process.}},
  Vol.~\bibinfo{volume}{28}. \bibinfo{publisher}{IEEE},
  \bibinfo{pages}{5881--5896}.
\newblock


\bibitem[\protect\citeauthoryear{Li, Zhang, Lu, and Shechtman}{Li
  et~al\mbox{.}}{2020}]%
        {li2020few}
\bibfield{author}{\bibinfo{person}{Yijun Li}, \bibinfo{person}{Richard Zhang},
  \bibinfo{person}{Jingwan Lu}, {and} \bibinfo{person}{Eli Shechtman}.}
  \bibinfo{year}{2020}\natexlab{}.
\newblock \showarticletitle{Few-shot Image Generation with Elastic Weight
  Consolidation}. In \bibinfo{booktitle}{\emph{Proc. NeurIPS}}.
\newblock


\bibitem[\protect\citeauthoryear{Liang, Liu, and Liu}{Liang
  et~al\mbox{.}}{2020}]%
        {liang2020dawson}
\bibfield{author}{\bibinfo{person}{Weixin Liang}, \bibinfo{person}{Zixuan Liu},
  {and} \bibinfo{person}{Can Liu}.} \bibinfo{year}{2020}\natexlab{}.
\newblock \showarticletitle{DAWSON: A Domain Adaptive Few Shot Generation
  Framework}. In \bibinfo{booktitle}{\emph{arXiv preprint arXiv:2001.00576}}.
\newblock


\bibitem[\protect\citeauthoryear{Mangla, Kumari, Singh, Balasubramanian, and
  Krishnamurthy}{Mangla et~al\mbox{.}}{2020}]%
        {mangla2020data}
\bibfield{author}{\bibinfo{person}{Puneet Mangla}, \bibinfo{person}{Nupur
  Kumari}, \bibinfo{person}{Mayank Singh}, \bibinfo{person}{Vineeth~N
  Balasubramanian}, {and} \bibinfo{person}{Balaji Krishnamurthy}.}
  \bibinfo{year}{2020}\natexlab{}.
\newblock \showarticletitle{Data Instance Prior for Transfer Learning in GANs}.
  In \bibinfo{booktitle}{\emph{arXiv preprint arXiv:2012.04256}}.
\newblock


\bibitem[\protect\citeauthoryear{Mescheder, Geiger, and Nowozin}{Mescheder
  et~al\mbox{.}}{2018}]%
        {mescheder2018training}
\bibfield{author}{\bibinfo{person}{Lars Mescheder}, \bibinfo{person}{Andreas
  Geiger}, {and} \bibinfo{person}{Sebastian Nowozin}.}
  \bibinfo{year}{2018}\natexlab{}.
\newblock \showarticletitle{Which training methods for GANs do actually
  converge?}. In \bibinfo{booktitle}{\emph{Proc. ICML}}. PMLR,
  \bibinfo{pages}{3481--3490}.
\newblock


\bibitem[\protect\citeauthoryear{Miyato, Kataoka, Koyama, and Yoshida}{Miyato
  et~al\mbox{.}}{2018}]%
        {miyato2018spectral}
\bibfield{author}{\bibinfo{person}{Takeru Miyato}, \bibinfo{person}{Toshiki
  Kataoka}, \bibinfo{person}{Masanori Koyama}, {and} \bibinfo{person}{Yuichi
  Yoshida}.} \bibinfo{year}{2018}\natexlab{}.
\newblock \showarticletitle{Spectral normalization for generative adversarial
  networks}. In \bibinfo{booktitle}{\emph{Proc. ICLR}}.
\newblock


\bibitem[\protect\citeauthoryear{Mo, Cho, and Shin}{Mo et~al\mbox{.}}{2020}]%
        {mo2020freeze}
\bibfield{author}{\bibinfo{person}{Sangwoo Mo}, \bibinfo{person}{Minsu Cho},
  {and} \bibinfo{person}{Jinwoo Shin}.} \bibinfo{year}{2020}\natexlab{}.
\newblock \showarticletitle{Freeze Discriminator: A Simple Baseline for
  Fine-tuning GANs}. In \bibinfo{booktitle}{\emph{Proc. CVPR Workshop}}.
\newblock


\bibitem[\protect\citeauthoryear{Nichol, Achiam, and Schulman}{Nichol
  et~al\mbox{.}}{2018}]%
        {nichol2018first}
\bibfield{author}{\bibinfo{person}{Alex Nichol}, \bibinfo{person}{Joshua
  Achiam}, {and} \bibinfo{person}{John Schulman}.}
  \bibinfo{year}{2018}\natexlab{}.
\newblock \showarticletitle{On first-order meta-learning algorithms}. In
  \bibinfo{booktitle}{\emph{arXiv preprint arXiv:1803.02999}}.
\newblock


\bibitem[\protect\citeauthoryear{Noguchi and Harada}{Noguchi and
  Harada}{2019}]%
        {noguchi2019image}
\bibfield{author}{\bibinfo{person}{Atsuhiro Noguchi} {and}
  \bibinfo{person}{Tatsuya Harada}.} \bibinfo{year}{2019}\natexlab{}.
\newblock \showarticletitle{Image generation from small datasets via batch
  statistics adaptation}. In \bibinfo{booktitle}{\emph{Proc. ICCV}}.
  \bibinfo{pages}{2750--2758}.
\newblock


\bibitem[\protect\citeauthoryear{Nowozin, Cseke, and Tomioka}{Nowozin
  et~al\mbox{.}}{2016}]%
        {nowozin2016f}
\bibfield{author}{\bibinfo{person}{Sebastian Nowozin}, \bibinfo{person}{Botond
  Cseke}, {and} \bibinfo{person}{Ryota Tomioka}.}
  \bibinfo{year}{2016}\natexlab{}.
\newblock \showarticletitle{f-gan: Training generative neural samplers using
  variational divergence minimization}. In \bibinfo{booktitle}{\emph{Proc.
  NeurIPS}}.
\newblock


\bibitem[\protect\citeauthoryear{Phaphuangwittayakul, Guo, and
  Ying}{Phaphuangwittayakul et~al\mbox{.}}{2021}]%
        {phaphuangwittayakul2021fast}
\bibfield{author}{\bibinfo{person}{Aniwat Phaphuangwittayakul},
  \bibinfo{person}{Yi Guo}, {and} \bibinfo{person}{Fangli Ying}.}
  \bibinfo{year}{2021}\natexlab{}.
\newblock \showarticletitle{Fast Adaptive Meta-Learning for Few-shot Image
  Generation}. In \bibinfo{booktitle}{\emph{{IEEE} Trans. Multimedia}}.
  \bibinfo{publisher}{IEEE}.
\newblock


\bibitem[\protect\citeauthoryear{Radford, Metz, and Chintala}{Radford
  et~al\mbox{.}}{2016}]%
        {radford2015unsupervised}
\bibfield{author}{\bibinfo{person}{Alec Radford}, \bibinfo{person}{Luke Metz},
  {and} \bibinfo{person}{Soumith Chintala}.} \bibinfo{year}{2016}\natexlab{}.
\newblock \showarticletitle{Unsupervised representation learning with deep
  convolutional generative adversarial networks}. In
  \bibinfo{booktitle}{\emph{Proc. ICLR}}.
\newblock


\bibitem[\protect\citeauthoryear{Robb, Chu, Kumar, and Huang}{Robb
  et~al\mbox{.}}{2020}]%
        {robb2020few}
\bibfield{author}{\bibinfo{person}{Esther Robb}, \bibinfo{person}{Wen-Sheng
  Chu}, \bibinfo{person}{Abhishek Kumar}, {and} \bibinfo{person}{Jia-Bin
  Huang}.} \bibinfo{year}{2020}\natexlab{}.
\newblock \showarticletitle{Few-Shot Adaptation of Generative Adversarial
  Networks}. In \bibinfo{booktitle}{\emph{arXiv preprint arXiv:2010.11943}}.
\newblock


\bibitem[\protect\citeauthoryear{Si and Zhu}{Si and Zhu}{2011}]%
        {si2011learning}
\bibfield{author}{\bibinfo{person}{Zhangzhang Si} {and}
  \bibinfo{person}{Song-Chun Zhu}.} \bibinfo{year}{2011}\natexlab{}.
\newblock \showarticletitle{Learning hybrid image templates (HIT) by
  information projection}. In \bibinfo{booktitle}{\emph{{IEEE} Trans. Pattern
  Anal. Mach. Intell.}}, Vol.~\bibinfo{volume}{34}. \bibinfo{publisher}{IEEE},
  \bibinfo{pages}{1354--1367}.
\newblock


\bibitem[\protect\citeauthoryear{Szegedy, Vanhoucke, Ioffe, Shlens, and
  Wojna}{Szegedy et~al\mbox{.}}{2016}]%
        {szegedy2016rethinking}
\bibfield{author}{\bibinfo{person}{Christian Szegedy}, \bibinfo{person}{Vincent
  Vanhoucke}, \bibinfo{person}{Sergey Ioffe}, \bibinfo{person}{Jon Shlens},
  {and} \bibinfo{person}{Zbigniew Wojna}.} \bibinfo{year}{2016}\natexlab{}.
\newblock \showarticletitle{Rethinking the inception architecture for computer
  vision}. In \bibinfo{booktitle}{\emph{Proc. CVPR}}.
  \bibinfo{pages}{2818--2826}.
\newblock


\bibitem[\protect\citeauthoryear{Tang, Liu, and Sebe}{Tang
  et~al\mbox{.}}{2020}]%
        {tang2020unified}
\bibfield{author}{\bibinfo{person}{Hao Tang}, \bibinfo{person}{Hong Liu}, {and}
  \bibinfo{person}{Nicu Sebe}.} \bibinfo{year}{2020}\natexlab{}.
\newblock \showarticletitle{Unified generative adversarial networks for
  controllable image-to-image translation}. In \bibinfo{booktitle}{\emph{{IEEE}
  Trans. Image Process.}}, Vol.~\bibinfo{volume}{29}.
  \bibinfo{publisher}{IEEE}, \bibinfo{pages}{8916--8929}.
\newblock


\bibitem[\protect\citeauthoryear{Tran, Tran, Nguyen, Nguyen, and Cheung}{Tran
  et~al\mbox{.}}{2021}]%
        {tran2021data}
\bibfield{author}{\bibinfo{person}{Ngoc-Trung Tran}, \bibinfo{person}{Viet-Hung
  Tran}, \bibinfo{person}{Ngoc-Bao Nguyen}, \bibinfo{person}{Trung-Kien
  Nguyen}, {and} \bibinfo{person}{Ngai-Man Cheung}.}
  \bibinfo{year}{2021}\natexlab{}.
\newblock \showarticletitle{On data augmentation for GAN training}. In
  \bibinfo{booktitle}{\emph{{IEEE} Trans. Image Process.}}
  \bibinfo{publisher}{IEEE}, \bibinfo{pages}{1882--1897}.
\newblock


\bibitem[\protect\citeauthoryear{Wang, Liu, Tao, Liu, Kautz, and
  Catanzaro}{Wang et~al\mbox{.}}{2019a}]%
        {wang2019fewshotvid2vid}
\bibfield{author}{\bibinfo{person}{Ting-Chun Wang}, \bibinfo{person}{Ming-Yu
  Liu}, \bibinfo{person}{Andrew Tao}, \bibinfo{person}{Guilin Liu},
  \bibinfo{person}{Jan Kautz}, {and} \bibinfo{person}{Bryan Catanzaro}.}
  \bibinfo{year}{2019}\natexlab{a}.
\newblock \showarticletitle{Few-shot Video-to-Video Synthesis}. In
  \bibinfo{booktitle}{\emph{Proc. NeurIPS}}.
\newblock


\bibitem[\protect\citeauthoryear{Wang, Wang, and Li}{Wang
  et~al\mbox{.}}{2019b}]%
        {wang2019u}
\bibfield{author}{\bibinfo{person}{Xueping Wang}, \bibinfo{person}{Yunhong
  Wang}, {and} \bibinfo{person}{Weixin Li}.} \bibinfo{year}{2019}\natexlab{b}.
\newblock \showarticletitle{U-Net conditional GANs for photo-realistic and
  identity-preserving facial expression synthesis}. In
  \bibinfo{booktitle}{\emph{ACM Trans. Multimedia Comput. Commun. Appl.}},
  Vol.~\bibinfo{volume}{15}. \bibinfo{publisher}{ACM New York, NY, USA},
  \bibinfo{pages}{1--23}.
\newblock


\bibitem[\protect\citeauthoryear{Wang, Gonzalez-Garcia, Berga, Herranz, Khan,
  and Weijer}{Wang et~al\mbox{.}}{2020}]%
        {wang2020minegan}
\bibfield{author}{\bibinfo{person}{Yaxing Wang}, \bibinfo{person}{Abel
  Gonzalez-Garcia}, \bibinfo{person}{David Berga}, \bibinfo{person}{Luis
  Herranz}, \bibinfo{person}{Fahad~Shahbaz Khan}, {and} \bibinfo{person}{Joost
  van~de Weijer}.} \bibinfo{year}{2020}\natexlab{}.
\newblock \showarticletitle{Minegan: effective knowledge transfer from gans to
  target domains with few images}. In \bibinfo{booktitle}{\emph{Proc. CVPR}}.
  \bibinfo{pages}{9332--9341}.
\newblock


\bibitem[\protect\citeauthoryear{Wang, Wu, Herranz, van~de Weijer,
  Gonzalez-Garcia, and Raducanu}{Wang et~al\mbox{.}}{2018}]%
        {wang2018transferring}
\bibfield{author}{\bibinfo{person}{Yaxing Wang}, \bibinfo{person}{Chenshen Wu},
  \bibinfo{person}{Luis Herranz}, \bibinfo{person}{Joost van~de Weijer},
  \bibinfo{person}{Abel Gonzalez-Garcia}, {and} \bibinfo{person}{Bogdan
  Raducanu}.} \bibinfo{year}{2018}\natexlab{}.
\newblock \showarticletitle{Transferring gans: generating images from limited
  data}. In \bibinfo{booktitle}{\emph{Proc. ECCV}}. \bibinfo{pages}{218--234}.
\newblock


\bibitem[\protect\citeauthoryear{Wang and Yao}{Wang and Yao}{2019}]%
        {wang2019few}
\bibfield{author}{\bibinfo{person}{Yaqing Wang} {and} \bibinfo{person}{Quanming
  Yao}.} \bibinfo{year}{2019}\natexlab{}.
\newblock \showarticletitle{Few-shot learning: A survey}. In
  \bibinfo{booktitle}{\emph{arXiv preprint arXiv:1904.05046}}.
\newblock


\bibitem[\protect\citeauthoryear{Yang, Shen, Xu, and Zhou}{Yang
  et~al\mbox{.}}{2021}]%
        {yang2021data}
\bibfield{author}{\bibinfo{person}{Ceyuan Yang}, \bibinfo{person}{Yujun Shen},
  \bibinfo{person}{Yinghao Xu}, {and} \bibinfo{person}{Bolei Zhou}.}
  \bibinfo{year}{2021}\natexlab{}.
\newblock \showarticletitle{Data-Efficient Instance Generation from Instance
  Discrimination}. In \bibinfo{booktitle}{\emph{arXiv preprint
  arXiv:2106.04566}}.
\newblock


\bibitem[\protect\citeauthoryear{Yu, Zhang, Song, Seff, and Xiao}{Yu
  et~al\mbox{.}}{2015}]%
        {yu15lsun}
\bibfield{author}{\bibinfo{person}{Fisher Yu}, \bibinfo{person}{Yinda Zhang},
  \bibinfo{person}{Shuran Song}, \bibinfo{person}{Ari Seff}, {and}
  \bibinfo{person}{Jianxiong Xiao}.} \bibinfo{year}{2015}\natexlab{}.
\newblock \showarticletitle{LSUN: Construction of a Large-scale Image Dataset
  using Deep Learning with Humans in the Loop}. In
  \bibinfo{booktitle}{\emph{arXiv preprint arXiv:1506.03365}}.
\newblock


\bibitem[\protect\citeauthoryear{Zhang, Goodfellow, Metaxas, and Odena}{Zhang
  et~al\mbox{.}}{2019a}]%
        {zhang2019self}
\bibfield{author}{\bibinfo{person}{Han Zhang}, \bibinfo{person}{Ian
  Goodfellow}, \bibinfo{person}{Dimitris Metaxas}, {and}
  \bibinfo{person}{Augustus Odena}.} \bibinfo{year}{2019}\natexlab{a}.
\newblock \showarticletitle{Self-attention generative adversarial networks}. In
  \bibinfo{booktitle}{\emph{Proc. ICML}}. PMLR, \bibinfo{pages}{7354--7363}.
\newblock


\bibitem[\protect\citeauthoryear{Zhang, Zhang, Odena, and Lee}{Zhang
  et~al\mbox{.}}{2019b}]%
        {zhang2019consistency}
\bibfield{author}{\bibinfo{person}{Han Zhang}, \bibinfo{person}{Zizhao Zhang},
  \bibinfo{person}{Augustus Odena}, {and} \bibinfo{person}{Honglak Lee}.}
  \bibinfo{year}{2019}\natexlab{b}.
\newblock \showarticletitle{Consistency regularization for generative
  adversarial networks}. In \bibinfo{booktitle}{\emph{Proc. ICLR}}.
\newblock


\bibitem[\protect\citeauthoryear{Zhao, Mathieu, and LeCun}{Zhao
  et~al\mbox{.}}{2017}]%
        {zhao2016energy}
\bibfield{author}{\bibinfo{person}{Junbo Zhao}, \bibinfo{person}{Michael
  Mathieu}, {and} \bibinfo{person}{Yann LeCun}.}
  \bibinfo{year}{2017}\natexlab{}.
\newblock \showarticletitle{Energy-based generative adversarial network}. In
  \bibinfo{booktitle}{\emph{Proc. ICLR}}.
\newblock


\bibitem[\protect\citeauthoryear{Zhao, Cong, and Carin}{Zhao
  et~al\mbox{.}}{2020a}]%
        {zhao2020leveraging}
\bibfield{author}{\bibinfo{person}{Miaoyun Zhao}, \bibinfo{person}{Yulai Cong},
  {and} \bibinfo{person}{Lawrence Carin}.} \bibinfo{year}{2020}\natexlab{a}.
\newblock \showarticletitle{On leveraging pretrained GANs for generation with
  limited data}. In \bibinfo{booktitle}{\emph{Proc. ICML}}. PMLR,
  \bibinfo{pages}{11340--11351}.
\newblock


\bibitem[\protect\citeauthoryear{Zhao, Liu, Lin, Zhu, and Han}{Zhao
  et~al\mbox{.}}{2020b}]%
        {zhao2020differentiable}
\bibfield{author}{\bibinfo{person}{Shengyu Zhao}, \bibinfo{person}{Zhijian
  Liu}, \bibinfo{person}{Ji Lin}, \bibinfo{person}{Jun-Yan Zhu}, {and}
  \bibinfo{person}{Song Han}.} \bibinfo{year}{2020}\natexlab{b}.
\newblock \showarticletitle{Differentiable augmentation for data-efficient gan
  training}. In \bibinfo{booktitle}{\emph{Proc. NeurIPS}}.
\newblock


\bibitem[\protect\citeauthoryear{Zhao, Singh, Lee, Zhang, Odena, and
  Zhang}{Zhao et~al\mbox{.}}{2021}]%
        {zhao2020improved}
\bibfield{author}{\bibinfo{person}{Zhengli Zhao}, \bibinfo{person}{Sameer
  Singh}, \bibinfo{person}{Honglak Lee}, \bibinfo{person}{Zizhao Zhang},
  \bibinfo{person}{Augustus Odena}, {and} \bibinfo{person}{Han Zhang}.}
  \bibinfo{year}{2021}\natexlab{}.
\newblock \showarticletitle{Improved consistency regularization for gans}. In
  \bibinfo{booktitle}{\emph{Proc. AAAI}}.
\newblock


\bibitem[\protect\citeauthoryear{Zhao, Zhang, Chen, Singh, and Zhang}{Zhao
  et~al\mbox{.}}{2020c}]%
        {zhao2020image}
\bibfield{author}{\bibinfo{person}{Zhengli Zhao}, \bibinfo{person}{Zizhao
  Zhang}, \bibinfo{person}{Ting Chen}, \bibinfo{person}{Sameer Singh}, {and}
  \bibinfo{person}{Han Zhang}.} \bibinfo{year}{2020}\natexlab{c}.
\newblock \showarticletitle{Image augmentations for GAN training}. In
  \bibinfo{booktitle}{\emph{arXiv preprint arXiv:2006.02595}}.
\newblock


\bibitem[\protect\citeauthoryear{Zhu, Shen, Zhao, and Zhou}{Zhu
  et~al\mbox{.}}{2020}]%
        {zhu2020domain}
\bibfield{author}{\bibinfo{person}{Jiapeng Zhu}, \bibinfo{person}{Yujun Shen},
  \bibinfo{person}{Deli Zhao}, {and} \bibinfo{person}{Bolei Zhou}.}
  \bibinfo{year}{2020}\natexlab{}.
\newblock \showarticletitle{In-domain gan inversion for real image editing}. In
  \bibinfo{booktitle}{\emph{Proc. ECCV}}. Springer, \bibinfo{pages}{592--608}.
\newblock


\bibitem[\protect\citeauthoryear{Zhu, Kr{\"a}henb{\"u}hl, Shechtman, and
  Efros}{Zhu et~al\mbox{.}}{2016}]%
        {zhu2016generative}
\bibfield{author}{\bibinfo{person}{Jun-Yan Zhu}, \bibinfo{person}{Philipp
  Kr{\"a}henb{\"u}hl}, \bibinfo{person}{Eli Shechtman}, {and}
  \bibinfo{person}{Alexei~A Efros}.} \bibinfo{year}{2016}\natexlab{}.
\newblock \showarticletitle{Generative visual manipulation on the natural image
  manifold}. In \bibinfo{booktitle}{\emph{Proc. ECCV}}. Springer,
  \bibinfo{pages}{597--613}.
\newblock


\bibitem[\protect\citeauthoryear{Zhu, Fan, Luo, Xu, and Yang}{Zhu
  et~al\mbox{.}}{2021}]%
        {zhu2021few}
\bibfield{author}{\bibinfo{person}{Linchao Zhu}, \bibinfo{person}{Hehe Fan},
  \bibinfo{person}{Yawei Luo}, \bibinfo{person}{Mingliang Xu}, {and}
  \bibinfo{person}{Yi Yang}.} \bibinfo{year}{2021}\natexlab{}.
\newblock \showarticletitle{Few-Shot Common-Object Reasoning Using
  Common-Centric Localization Network}. In \bibinfo{booktitle}{\emph{{IEEE}
  Trans. Image Process.}}, Vol.~\bibinfo{volume}{30}.
  \bibinfo{publisher}{IEEE}, \bibinfo{pages}{4253--4262}.
\newblock


\end{thebibliography}

\end{document}